\documentclass[compsoc,journal]{IEEEtran}
\IEEEoverridecommandlockouts

\usepackage[utf8]{inputenc} % allow utf-8 input
\usepackage[T1]{fontenc}    % use 8-bit T1 fonts
\usepackage{hyperref}     
\usepackage{url}            % simple URL typesetting
\usepackage[export]{adjustbox}

\usepackage[misc]{ifsym}
\usepackage{epsfig}
\usepackage{graphicx}
\usepackage{algorithm,algpseudocode}
\usepackage{subfig}
\usepackage{enumitem}
\usepackage{array}
\usepackage{ragged2e}
\usepackage{color}
\usepackage{tikz}
\usepackage[edges]{forest}
\definecolor{hidden-draw}{RGB}{20,68,106}
\definecolor{hidden-pink}{RGB}{255,245,247}
\title{Towards Explainable Artificial Intelligence (XAI): A Data Mining Perspective}

\begin{document}
\author{Haoyi Xiong, Xuhong Li, Xiaofei Zhang, Jiamin Chen, Xinhao Sun, Yuchen Li, Zeyi Sun, and Mengnan Du}
% Himabindu Lakkaraju

\IEEEtitleabstractindextext{%
\begin{abstract}
%Deep neural networks (DNNs) have been widely used to enable artificial intelligence (AI) in a wide range of data-intensive applications, such as content understanding \& generation, medical diagnosing \& intervention, autonomous driving, etc. It has become an emerging research area to interpret and verify the behaviors of DNNs, despite their over-parameterized architectures and decision-making pathways in a black-box manner. Extensive efforts have been done to either build up DNNs using interpretable models or explain the behaviors of DNNs in human-readable ways. While contributions in this area have been majorly summarized and reviewed from a perspective of algorithms and models, this work is dedicated to a new ``data-centric'' storyline: How data has been collected, processed, and analyzed to explain DNNs for AI, or namely explainable Artificial Intelligence (XAI). Particularly, we consider explaining  as a data mining process on the training and testing datasets in various modalities, such as tabular data, texts, images, graphs and etc. 

Given the complexity and lack of transparency in deep neural networks (DNNs), extensive efforts have been made to make these systems more interpretable or explain their behaviors in accessible terms. Unlike most reviews, which focus on algorithmic and model-centric perspectives, this work takes a ``data-centric'' view, examining how data collection, processing, and analysis contribute to explainable AI (XAI). We categorize existing work into three categories subject to their purposes: \emph{interpretations of deep models}, referring to feature attributions and reasoning processes that correlate data points with model outputs; \emph{influences of training data}, examining the impact of training data nuances, such as data valuation and sample anomalies, on decision-making processes; and \emph{insights of domain knowledge}, discovering latent patterns and fostering new knowledge from data and models to advance social values and scientific discovery. Specifically, we distill XAI methodologies into data mining operations on training and testing data across modalities, such as images, text, and tabular data, as well as on training logs, checkpoints, models and other DNN behavior descriptors. In this way, our study offers a comprehensive, data-centric examination of XAI from a lens of data mining methods and applications.
\end{abstract}

}
\maketitle

\begin{figure*}
    \centering
    \includegraphics[width=0.95\textwidth]{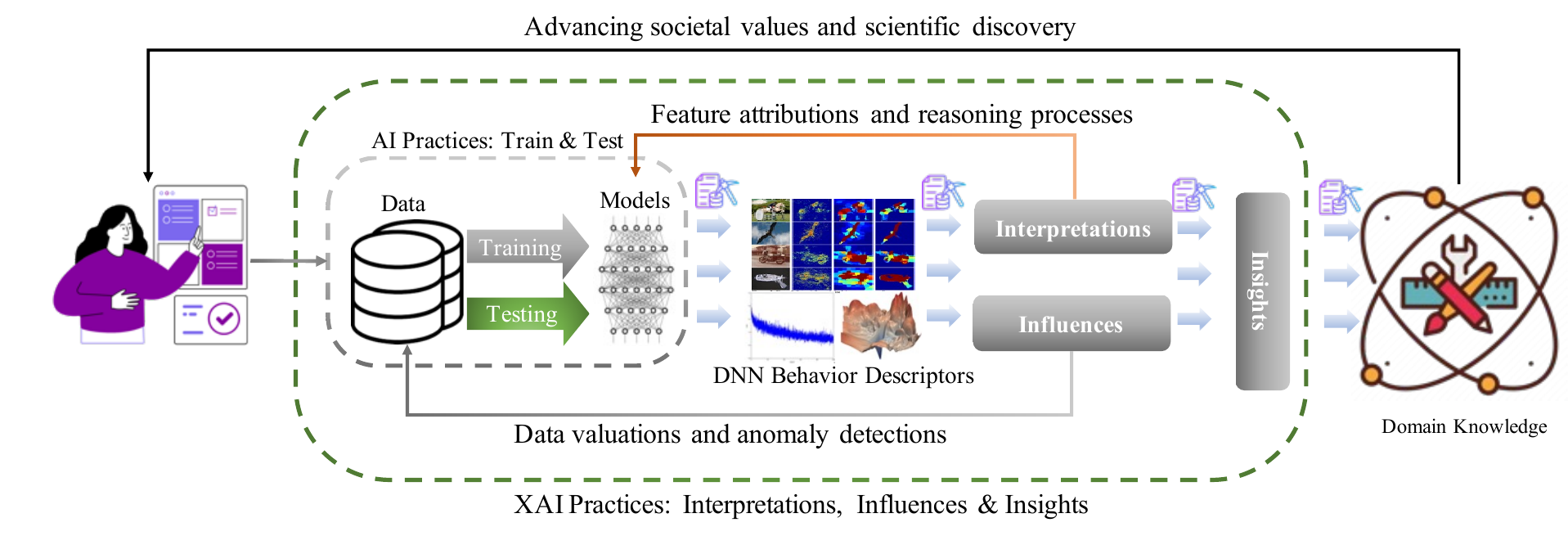}\vspace{-3mm}
    \caption{Overview of Explainable AI as a Data Mining Approach for Interpretations, Influences and Insights}
    \vspace{-5mm}
    \label{fig:data-mining-flow}
\end{figure*}

\section{Introduction}
As Artificial Intelligence (AI) has progressed, traditional decision-making techniques such as perceptron \cite{rosenblatt1958perceptron}, rule-based systems \cite{hayes1985rule}, case-based reasoning \cite{kolodner1992introduction}, and expert systems \cite{waterman1985guide} have given way to more complex DNNs \cite{lecun2015deep}. These early techniques were based on human decision-making processes, ranging from rule-based reasoning \cite{frye1995theory} to committee-based predictions \cite{dutta1997strategies}. The surge in storage and computational power has catalyzed this evolution towards DNNs, which despite their performance on tasks such as visual recognition and language modeling \cite{lecun2015deep}, face challenges in explainability \cite{doshi2017towards}.

The ``black-box'' nature of DNNs alongside their extensive parameterization impedes the transparency required in critical applications like autonomous driving and healthcare, sparking concern over the reliability of such models in high-stakes environments \cite{carvalho2019machine,jing2022inaction,wang2023interpretable}. As a result, explainable AI (XAI) has emerged as a crucial field, proposing solutions such as LIME \cite{ribeiro2016should} to improve the interpretability\footnote{In this work, the terms ``explainability" and ``interpretability," as well as ``explainable" and ``interpretable," are used interchangeably.} of machine learning, potentially increasing trust in AI systems \cite{toreini2020relationship}. These XAI techniques not only strive for model transparency but also contribute added value to datasets, helping with tasks such as debugging \cite{afzal2021data} and localizing mislabeled samples \cite{jia2023learning}, enriching the understanding of datasets and their respective domains \cite{marcinkevivcs2023interpretable,wang2023interpretable}. In this study, we thoroughly examine the available literature, grouping and analyzing them through the lens of our two unique observations, three purposes and four-stage data processes of XAI techniques, as follows.

%\subsection{Three Purposes of XAI Techniques}
Our first observation focuses on the driving forces behind the evolution and application of XAI techniques. Following an extensive review of the current literature in this domain, we have distilled the primary purposes into three core categories:  

\begin{enumerate}
 \item \textbf{Interpretations of Deep Models}: Despite the high predictive capabilities of deep learning models, their ``black-box'' nature limits interpretability \cite{ribeiro2016should,selvaraju2017grad}. XAI aims to illuminate these models by elucidating their prediction rationales on a per-instance basis, subsequently fostering transparency and trust \cite{doshi2017towards,zhang2018training}.

 \item \textbf{Influences of Training Data}: The performance of machine learning models is dependent on the distribution and quality of the training data~\cite{koh2017understanding,feldman2020neural}. XAI techniques can pinpoint data points that significantly affect model outputs, facilitating an improved training process and model simplicity \cite{datta2016algorithmic,NEURIPS2022_d0702278}.

 \item \textbf{Insights of Domain Knowledge}: XAI also sheds light on domain-specific knowledge within models and data, offering potential advances in human understanding within these domains and proving valuable insights in high-stakes applications such as healthcare and finance \cite{ghosh2020visualisation,zednik2022scientific}.

\end{enumerate}
As shown in Fig.~\ref{fig:data-mining-flow}, XAI serves as a bridge for the gap between human comprehension and the complexity of machine learning models, improving confidence in AI applications \cite{li2022interpretable,nauta2023interpreting}.

%\subsection{XAI as Data Mining approaches}
We also found that the XAI methodologies follow a structured process similar to traditional data mining \cite{chen1996data,wu2013data,han2022data}, integrating data, algorithms, and human-centric analyses. The four integral steps are outlined below.
\begin{enumerate}
    \item \textbf{Data Acquisition \& Collection}: XAI extends data collection beyond datasets but covers the life-cycle of deep learning, such as training datasets, training logs and checkpoints, testing samples, and so on.
    
    \item \textbf{Data Preparation \& Transformation}: The behavior descriptors of DNNs, including saliency maps, training loss curves, and input/loss gradient vectors (please see also Table~\ref{tab:behavior_descriptors}), are extracted and transformed from models, data and training logs to allow subsequent explanations \cite{tjoa2020survey,feng2021phases,jia2023learning}.
        
    \item \textbf{Data Modeling \& Analyses}: The mining of DNN behavior descriptors serves to model DNN decisions, training data contributions, and dataset patterns, leading to three types of analytical purposes: interpretations, influences, and insights \cite{wang2023interpretable}.
    
    \item \textbf{Results Reporting \& Visualizations}: The culmination of XAI efforts is the presentation of findings via appropriate reporting and visualizations subject to the data modality, such as overlaying saliency maps onto images \cite{qi2019visualizing,wang2020cnn} to highlight the key visual features attributed.

\end{enumerate}
Through these steps, XAI enhances the interpretability, trust, and even knowledge \& understanding in AI frameworks, fostering improved human-AI synergy.

%\subsection{Our Work}
Our investigation adopts a data-centric perspective for examining XAI, categorically organizing techniques by combining \emph{three purposes} with \emph{four-stage data mining processes}. The contributions of this study include:
%In summary, we make the following contributions.
\begin{itemize}
    \item A technical review of the XAI paradigm from a data mining perspective, focusing on data-related practices in the explanation processes \cite{moher2009preferred}. This work pioneers the systematic examination of XAI in a new framework.
    
    \item The introduction of a novel \emph{taxonomy system}, structured around both the \textbf{threefold purposes} of XAI and the \textbf{four distinct stages} of data mining, to categorize and elucidate current XAI methodologies.
    
    \item A forward-looking discourse on the prospective evolution of XAI, emphasizing its capacity to uncover deeper insights within data, which has significant implications for domains like AI-driven science and medicine.
    
\end{itemize}
The organization of XAI studies within this taxonomy provides a structured narrative, enriching the discourse with a precise understanding of the trends and potential in XAI.

\tikzstyle{my-box}=[
    rectangle,
    draw=hidden-draw,
    rounded corners,
    text opacity=1,
    minimum height=1.5em,
    minimum width=5em,
    inner sep=2pt,
    align=center,
    fill opacity=.5,
    line width=0.8pt,
]
\tikzstyle{leaf}=[my-box, minimum height=1.5em,
    fill=hidden-pink!80, text=black, align=left,font=\normalsize,
    inner xsep=2pt,
    inner ysep=4pt,
    line width=0.8pt,
]
\begin{figure*}[t!]
    \centering
    \resizebox{\textwidth}{!}{
        \begin{forest}
            forked edges,
            for tree={
                grow=east,
                reversed=true,
                anchor=base west,
                parent anchor=east,
                child anchor=west,
                base=center,
                font=\large,
                rectangle,
                draw=hidden-draw,
                rounded corners,
                align=left,
                text centered,
                minimum width=4em,
                edge+={darkgray, line width=1pt},
                s sep=3pt,
                inner xsep=2pt,
                inner ysep=3pt,
                line width=0.8pt,
                ver/.style={rotate=90, child anchor=north, parent anchor=south, anchor=center},
            },
            where level=1{text width=8em,font=\normalsize,}{},
            where level=2{text width=10em,font=\normalsize,}{},
            where level=3{text width=12em,font=\normalsize,}{},
            where level=4{text width=7em,font=\normalsize,}{},
            [
                Towards Explainable Artificial Intelligence (XAI): A Data Mining Perspective, ver
                [
                    Interpretations
                    [
                        % Instructions From \\ Human  (\S \ref{instructionfromhuman})
                        Feature Attributions\\as Model Explanation
                        [
                            Perturbation-based\\ Methods
                            [                            
                                LIME~\cite{ribeiro2016should}{, }G-LIME~\cite{li2023g}{, } Feature Ablation~\cite{merrick2019randomized,ramaswamy2020ablation}{, } SHAP~\cite{lundberg2017unified}{, }\\  BSHAP~\cite{sundararajan2020many}{, } DI Shapley~\cite{aas2021explaining}{, }
                                MV Shapley~\cite{zhang2021interpreting}{,} Shapley-Taylor~\cite{sundararajan2020shapley}{,} 
                                , leaf, text width=32em
                            ]
                        ]
                        [
                            Differentiation-based\\ Methods
                            [
                                Integrated Gradients~\cite{qi2019visualizing,lundstrom2022rigorous}{, } SmoothGrad~\cite{smilkov2017smoothgrad}{, }
                                DeepLIFT~\cite{shrikumar2017learning}{, }\\DeepSHAP~\cite{fernando2019study}{, }
                                Grad-CAM~\cite{selvaraju2017grad}{, }
                                GradSHAP~\cite{lundberg2017unified}{, }
                                , leaf, text width=32em
                            ]
                        ]
                        [
                            Activation \& Attention-\\based Methods
                            [
                                CAM~\cite{zhou2016learning}{, }
                                Grad-CAM~\cite{selvaraju2017grad}{, } Attention Flow~\cite{abnar2020quantifying}{, }
                                LRP~\cite{voita2019analyzing}{, }\\Attention Rollout~\cite{xu2023attribution}{, }
                                Bidirectional Attention~\cite{chen2022beyond}
                                , leaf, text width=32em
                            ]
                        ]
                        [
                            Proxy Explainable\\ Models
                            [
                                LIME~\cite{ribeiro2016should}{, }G-LIME~\cite{li2023g}{, }Distilled Soft Decision Trees~\cite{frosst2017distilling}{, }Surro-\\
                                gate Decision Trees~\cite{di2019surrogate}{, }Deep Neural Decision Trees~\cite{kontschieder2015deep}{, }DTD~\cite{montavon2017explaining} 
                                , leaf, text width=32em
                            ]
                        ]
                    ]
                    [
                        % Instruction Data \\ Management (\S \ref{datamanagement})
                        Reasoning Process as\\ Model Explanation
                        [
                            Visualizing Intermediate\\ Representation
                            [
                                Network Dissection~\cite{bau2017network,zhou2018interpreting}{, } DeepVis~\cite{yosinski2015understanding}{, } DeepDream~\cite{couteaux2019towards}\\
                                Eigen-CAM~\cite{muhammad2020eigen}{, }  Multifaceted Feature Visualization~\cite{quach2023using,zeiler2014visualizing}{, }
                                , leaf, text width=32em
                            ]
                        ]
                        [
                            Visualizing the Logic of \\Reasoning
                            [
                                Tree Models Extraction~\cite{boz2002extracting}{, }Distilled Soft Decision Trees~\cite{frosst2017distilling}{, } \\
                                Distilled Gradient Boosting Trees~\cite{zhangquanshi2019interpreting}{, }
                                , leaf, text width=32em
                            ]
                        ]
                        [
                            Counterfactual Examples \\as Decision Rules
                            [
                                FIDO~\cite{chang2018explaining}{, }DiCE~\cite{mothilal2020explaining}{, }Counterfactual Visual Explanations~\cite{goyal2019counterfactual}{, }\\Laugel~\emph{et al.}~\cite{laugel2019unjustified}{, }Ilyas~\emph{et al.}~\cite{ilyas2019adversarial}
                                , leaf, text width=32em
                            ]
                        ]
                        [
                            Prototypes as Decision\\ Rules
                            [
                                ProtoPNet~\cite{chen2019looks}{, }Deformable ProtoPNet~\cite{donnelly2022deformable}{, }PIP-Net~\cite{nauta2023pip}{, }Interpret-\\Net~\cite{singh2022think}{, }Co-12~\cite{nauta2023co}{, }HQProtoPNet~\cite{wang2023hqprotopnet}{, }Prototype Selection~\cite{bien2011prototype}
                                , leaf, text width=32em
                            ]
                        ]
                        [
                            Concept Activation Vectors\\ and Derivatives
                            [
                                CAV and TCAV~\cite{kim2018interpretability}{, }Invertible CAV~\cite{zhang2021invertible}{, }Text2Concept~\cite{moayeri2023text2concept}{, }\\Concept Activation Regions~\cite{crabbe2022concept}{, }
                                , leaf, text width=32em
                            ]
                        ]
                    ]
                ]
                [
                    Influences
                    [
                        % Instructions From \\ Human  (\S \ref{instructionfromhuman})
                        Sample Valuation as\\ Model Explanation
                        [
                            Gradient-based Methods\\ for Valuation
                            [                            
                                Influence Functions~\cite{koh2017understanding}{, }TracIn~\cite{NEURIPS2020_e6385d39}{, }FastIF~\cite{guo2021fastif}{, }Bhatt et al.~\cite{bhatt2021fast}\\Grosse~\emph{et al.}~\cite{grosse2023studying}{, }
                                Liu~\emph{et al.}~\cite{liu2022debugging}{, }Wang~\emph{et al}~\cite{wang2020less}{, }Bae~\emph{et al.}~\cite{bae2022if}{, }\\Cross-Loss IF~\cite{silva2022cross}
                                , leaf, text width=32em
                            ]
                        ]
                        [
                            Resampling-based\\ Methods for Valuation
                            [
                                Leave-One-Out (LOO)~\cite{bates2023cross}{, }Jackknife~\cite{efron1992jackknife}{, }
                                % Smirnov et al.~\cite{smirnov2018hard}{, } O2u-Net~\cite{huang2019o2u}{, }
                                % Jia et al.~\cite{jia2023learning}{, }\\ TAPUDD~\cite{dua2023task}{, }
                                % PaLM~\cite{krishnan2017palm}{,}
                                , leaf, text width=32em
                            ]
                        ]
                        [
                            Game theoretical\\ Methods for Valuation
                            [
                                Data Shapley~\cite{ghorbani2019data}{, }Beta Shapley~\cite{kwon2022beta}{, }Data Banzaf~\cite{wang2023data}{, }DataInf~\cite{kwon2023datainf}\\CS-Shapley~\cite{schoch2022cs}{, }LAVA~\cite{just2022lava}
                                % Dolly-v2~\cite{DatabricksBlog2023DollyV2}{, } OpenAssistant~\cite{Kopf2023OpenAssistantC}{, } \\
                                % COIG~\cite{Zhang2023ChineseOI}{,} ShareGPT~\cite{vicuna2023}{,} 
                                , leaf, text width=32em
                            ]
                        ]
                    ]
                    [
                        % Instruction Data \\ Management (\S \ref{datamanagement})
                        Sample Anomalies as\\ Model Explanation
                        [
                            Hard Sample Mining
                            [
                                HEM~\cite{zhou2023samples}{, }Auxiliary Embeddings~\cite{smirnov2018hard}{, }Detection of Forgetting\\ Events~\cite{toneva2018empirical}{, }
                                Shrivastava~\emph{et al.}~\cite{shrivastava2016training}{, }
                                Focal Loss~\cite{lin2017focal}{, }
                                Smirnov~\emph{et al.}~\cite{smirnov2018hard}{, }\\
                                Triplet Loss \cite{chechik2010large}{}
                                , leaf, text width=32em
                            ]
                        ]
                        [
                            Mislabel Detection
                            [
                                Visual and Interactive Analytics~\cite{xiang2019interactive,zhang2020dealing}{, }OOD~\cite{chen2020oodanalyzer}{, }O2u-Net~\cite{huang2019o2u}{, }\\Kernel methods~\cite{valizadegan2007kernel}{, }
                                Approach for Multi-label Learning tasks~\cite{xie2021partial}{, }\\
                                Mitigating the influence of mislabeled samples on the training\\ processes of deep learning models~\cite{feng2021phases,guan2013survey,gupta2019dealing,song2022learning}{, }
                                % Jia et al.~\cite{jia2023learning}{, }\\
                                % Human Oversight~\cite{valizadegan2007kernel,huang2019o2u,xie2021partial}{, }
                                , leaf, text width=32em
                            ]
                        ]
                        [
                            Dataset Debugging
                            [
                                Cinquini et al.~\cite{cinquini2023handling}{, }Sejr et al.~\cite{sejr2021explainable}{, }
                                KDE~\cite{cao2005goodness}{, }Kolmogorov-Smirnov \\Test~\cite{glazer2012learning}{, }
                                Martinez et al.~\cite{martinez2022can}{, }
                                TAPUDD~\cite{dua2023task}{, }
                                Holmberg et al.~\cite{holmberg2023exploring}{, }\\
                                De et al.~\cite{de2023weighted}{, }
                                Zhang et al.~\cite{zhang2018training}{, }PaLM~\cite{krishnan2017palm}{, }
                                MLInspect~\cite{grafberger2022data}{, }
                                , leaf, text width=32em
                            ]
                        ]
                    ]
                ]
                [
                    Insights
                    [
                        % Instructions From \\ Human  (\S \ref{instructionfromhuman})
                        Societal Values as\\ Model Explanation
                        [
                            Algorithmic Fairness
                            [                            
                                Madras et al.~\cite{madras2018learning}{, }
                                Zhou et al.~\cite{zhou2023explain}{, }
                                Smith et al.~\cite{smith2023many}{,}
                                Gardner et al.~\cite{gardner2023cross}\\
                                Benbouzid et al.~\cite{benbouzid2023fairness}{, }
                                Shrestha et al.~\cite{shrestha2023help}
                                Castelnovo et al.~\cite{castelnovo2021towards}{, }\\
                                Stanley et al.~\cite{stanley2022fairness}{, }
                                Alikhademi et al.~\cite{alikhademi2021can}{, }
                                , leaf, text width=32em
                            ]
                        ]
                        [
                            Digital Ethics
                            [
                                McDermid et al.~\cite{mcdermid2021artificial}{, }
                                Rakova et al.~\cite{rakova2023algorithms}{, }
                                Jakesch et al.~\cite{jakesch2022different}{, }\\
                                Mccradden et al.~\cite{mccradden2023s}{, }
                                Kasirzadeh et al.~\cite{kasirzadeh2021use}{,} Hawkins et al.~\cite{hawkins2023ethical}{,} 
                                , leaf, text width=32em
                            ]
                        ]
                        [
                            Systems Accountability
                            [
                                Structural Causal Models~\cite{kacianka2021designing}{, } Hutchinson et al.~\cite{hutchinson2021towards}{, }\\
                                Kroll et al.~\cite{kroll2021outlining}
                                Lima et al.~\cite{lima2022conflict}{,} Donia et al.~\cite{donia2022normative}{,} 
                                , leaf, text width=32em
                            ]
                        ]
                        [
                            Decision Transparency
                            [
                                Healthsheet~\cite{rostamzadeh2022healthsheet}{, }Schmude et al.~\cite{schmude2023impact}{, }
                                TILT~\cite{grunewald2021tilt}{, }
                                Education~\cite{marian2023algorithmic}{, }\\
                                Public Administration~\cite{lapostol2023algorithmic}{, }
                                , leaf, text width=32em
                            ]
                        ]
                    ]
                    [
                        % Instruction Data \\ Management (\S \ref{datamanagement})
                        Scientific Explorations\\ as Model Explanation
                        [
                            Pattern Recognition
                            [
                                Drug Discovery~\cite{jimenez2021coloring,proietti2023explainable}\\
                                % Concept Whitenin~\cite{proietti2023explainable}{, }\\
                                Re-conceptualization of Newton’s law of gravitation~\cite{lemos2023rediscovering}{, }
                                , leaf, text width=32em
                            ]
                        ]
                        [
                            Interdisciplinary\\ Collaboration
                            [
                                Protein Folding~\cite{jumper2021highly}{, }
                                Molecular Property Prediction~\cite{heberle2023xsmiles}{, }\\
                                Medical Imaging~\cite{el2021multilayer, essemlali2020understanding, li2022interpretable}{, }
                                CT Scans~\cite{ardila2019end}{, }\\
                                Personalized Treatment Plans~\cite{basu2011developing}{, }
                                Borys et al.~\cite{borys2023explainable}{,}\\
                                Deciphering tinnitus patient data~\cite{tarnowska2021explainable}{, }
                                , leaf, text width=32em
                            ]
                        ]
                        [
                            Uncovering Mechanisms
                            [
                               % Natural Systems~\cite{burkart2021survey,carleo2019machine}{, }
                                Quantum Physics~\cite{gross2017quantum,melko2019restricted}{, }Genomics~\cite{keyl2023single,park2020global}\\
                                Sahin et al.~\cite{sahin2021xai}{, }
                                Anguita et al.~\cite{anguita2020explainable}{, }
                                Geosciences~\cite{li2023explainable, rozemberczki2022shapley}
                                , leaf, text width=32em
                            ]
                        ]
                    ]
                ]
            ]
        \end{forest}}
    \caption{Taxonomy of research in Explainable Artificial Intelligence (XAI) from a Data Mining Perspectives: Interpretation of Deep Models, Influences of Training Samples, and Insights of Domain Knowledge.}
    \label{fig:taxonomy}
\end{figure*}
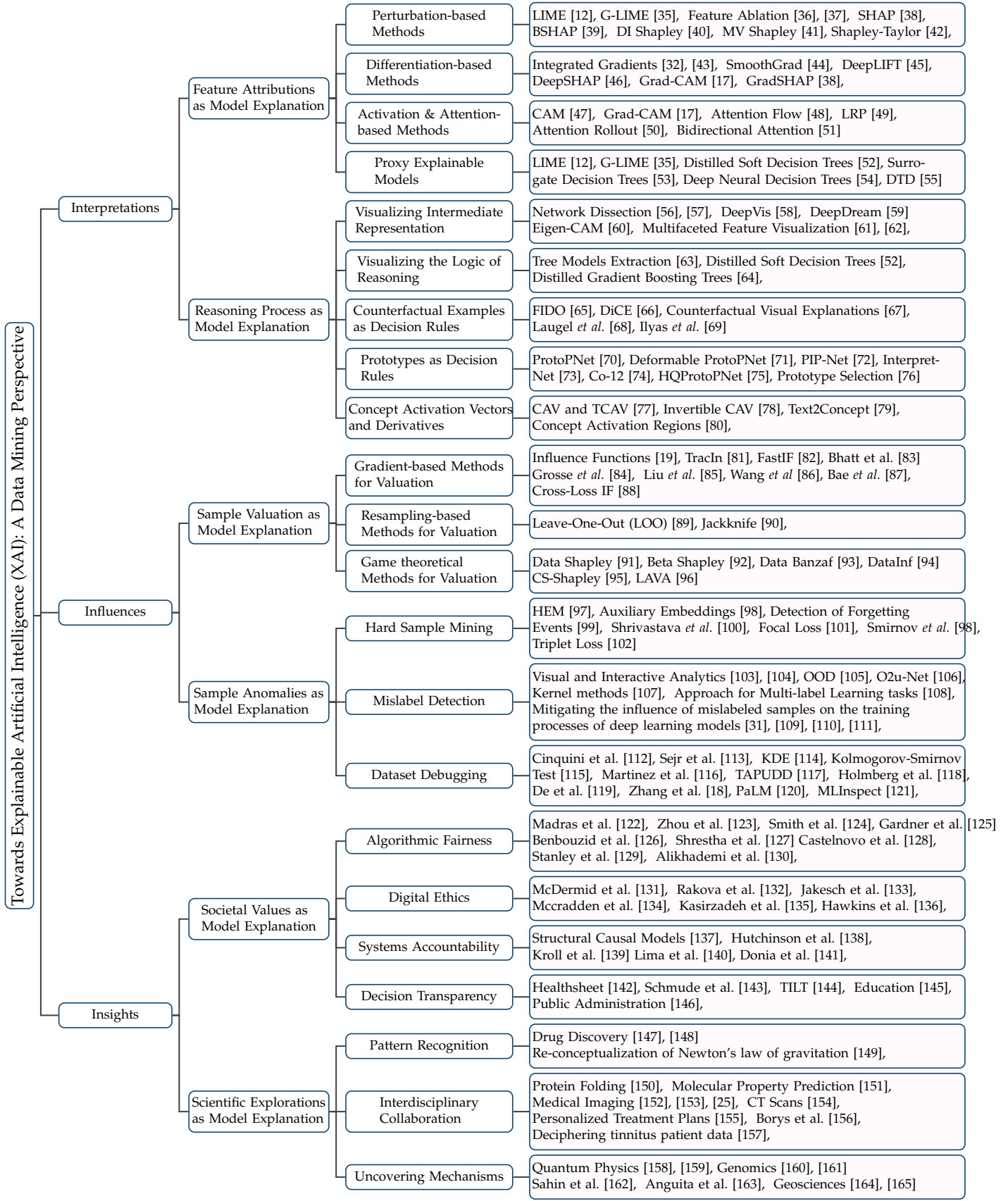

\begin{table*}[htbp] \caption{Some examples of DNN behavior descriptors, their data sources, and potential usages} \label{tab:behavior_descriptors} \vspace{-3mm}
\centering 
\begin{tabular}{|p{4cm}|p{6.5cm}|p{6.5cm}|} \hline 
\textbf{Name} & \textbf{Sources} & \textbf{Usages} \\ \hline 
Saliency Maps~\cite{simonyan2013deep,zeiler2014visualizing} & Extracting the activation values from various layers during the forward pass. & Insights into feature patterns and hierarchical feature extraction processes within the network. \\ \hline 
Layer Weights~\cite{martin2020heavy,martin2021implicit} & Analyzing the learned weights of neurons in each layer after the training process. & Understanding the generalization performance of DNN via random matrix theory. \\ \hline 
Training Loss Curve~\cite{feng2021phases,jia2023learning} & Monitoring and recording the loss (of every sample) during the model training process. & Assessing model performance, diagnosing overfitting or underfitting for every training sample. \\ \hline 
Gradients, Hessians~\cite{ancona2018towards,basu2020influence} & Computing gradients or hessians with respect to input or parameters using backpropagation. & Used for visual explanations, sensitivity analysis, and valuation of data influence. \\ \hline 
Attention Weights~\cite{yuan2021explaining,chen2022beyond} & Extracted from attention mechanisms in models like Transformers during inference. & Interpreting parts of input data (sequences) that the model focuses on to make predictions. \\ \hline  
Ablated Features~\cite{merrick2019randomized,ramaswamy2020ablation} & Systematic removal or alteration of features from the input and observing the change in model output. & Understanding the impact of individual features on model predictions and identifying critical features. \\ \hline 
Interpolated Samples~\cite{ribeiro2016should,li2023g} & Interpolating every sample through randomly perturbing the values of its features. & Understanding the impact of individual features on model predictions and identifying critical features. \\ \hline 
Adversarial Examples~\cite{ignatiev2019relating,geirhos2020shortcut} & Generating inputs specifically designed to cause the model to make incorrect predictions. & Identifying and addressing model vulnerabilities and improving the robustness of the network. \\ \hline 
\end{tabular} 
\vspace*{-5mm}
\end{table*}

Emerging research on XAI has been reviewed in several surveys, highlighting both the challenge and importance of interpreting deep models. Doshi-Velez and Kim~\cite{doshi2017towards} stress the necessity for evaluation techniques for XAI, while Carvalho \emph{et al.}~\cite{carvalho2019machine} offer an extensive study of interpretability methods covering model-agnostic and model-specific approaches. Hammoudeh and Lowd~\cite{hammoudeh2022training} shift the focus to the influence of training data. Mohseni~\emph{et al.} provide a survey and framework to evaluate XAI systems~\cite{mohseni2021multidisciplinary}. Practical XAI methods are expanded upon by Marcinkevi{\v{c}}s and Vogt~\cite{marcinkevivcs2023interpretable} and Notovich \emph{et al.}~\cite{notovich2023explainable}, providing application examples and a taxonomy of techniques. Domain-specific applications are explored by Preuer \emph{et al.}~\cite{preuer2019interpretable} in drug discovery and by Tjoa and Guan~\cite{tjoa2020survey} in medical imaging.

Compared to above work, our survey (a brief result shown in Fig.~\ref{fig:taxonomy}) bridges the gap in XAI literature by exploring the triple roles of XAI: (1) interpreting the behaviors of models to understand their decisions, (2) estimating the influences of data to valuate and identify key samples, and (3) distilling insights from both models and data to obtain new understandings that advance societal values and scientific discovery, all from a data mining perspective.

\section{Interpretations: Feature attributions and Reasoning Processes of Deep Models}
 Interpreting deep models involves using \emph{feature attribution} to assess the impact of each input on the model's output and examining \emph{reasoning processes} to understand the decision-making pathways within the model.

\subsection{Feature Attributions as Model Explanation}
To evaluate the significance of individual input features to predictions made by the model, 
some representative methods have been proposed as follows. 

\subsubsection{Perturbation-based Methods} 
Interpreting deep models through feature attributions is crucial in understanding their predictions, which is typically achieved via perturbation-based methods. These methods involve input modifications and assessing their effect on model output. A significant prediction change due to a small input perturbation indicates that the perturbed feature has high importance~\cite{noack2021empirical}.

Perturbation-based techniques like LIME~\cite{ribeiro2016should} construct local surrogate models to provide insight, though with computational demands and variability in results. G-LIME~\cite{li2023g} improves this by introducing a global prior-induced approach focusing on feature importance across datasets. Feature Ablation~\cite{merrick2019randomized,ramaswamy2020ablation} identifies critical features through systematic elimination but lacks consideration for feature interactions and is computationally expensive. Shapley values~\cite{shapley1953value} assess feature importance by computing their contribution to model predictions across all possible feature combinations~\cite{vstrumbelj2014explaining,datta2016algorithmic}. SHAP~\cite{lundberg2017unified} extends this concept to offer in-depth analysis, leading to more accurate and computationally efficient attribution variations such as BSHAP~\cite{sundararajan2020many}, and Shapley-Taylor~\cite{sundararajan2020shapley}.

Although these methods contribute to a more interpretable AI, they remain computationally intensive, particularly in high-dimensional spaces, posing a challenge for practical applications~\cite{rozemberczki2022shapley}.

\subsubsection{Differentiation-based Methods}  
 Differentiation-based methods form a significant branch in the interpretability of complex models via feature attributions. These techniques are rooted in the computation of gradients of a model's output with respect to its inputs, aligning each input feature with a corresponding gradient component that signifies its sensitivity—alterations in the feature cause identifiable shifts in model predictions~\cite{srinivas2020rethinking}.
 
 Integrated Gradients~\cite{qi2019visualizing,lundstrom2022rigorous} calculates the integral of the gradients along a path from a baseline to the input, capturing the importance of input features for the model's predictions. SmoothGrad~\cite{smilkov2017smoothgrad} enhances gradient-based interpretations by averaging the gradients of nearby points, reducing noise and refining the quality of attributions. DeepLIFT~\cite{shrikumar2017learning} advances transparency by contrasting the contributions of a feature against a reference point, differentiating predictive differences. DeepSHAP~\cite{fernando2019study} extends this by integrating Shapley values, comparing the effect of input features against a typical baseline, commonly a blank image in image recognition tasks. This integration enables a more nuanced evaluation of the importance of each input feature. 
 %Grad-CAM~\cite{selvaraju2017grad} utilizes gradients from convolutional layers to create visual explanations, illustrating pivotal image regions affecting model outcomes. 
 GradSHAP~\cite{lundberg2017unified} combines gradient signals and Shapley values to attribute feature significance, emphasizing influences on decision-making.

\begin{figure*}
    \centering
    \subfloat[ViT-B]{\includegraphics[width=0.48\textwidth]{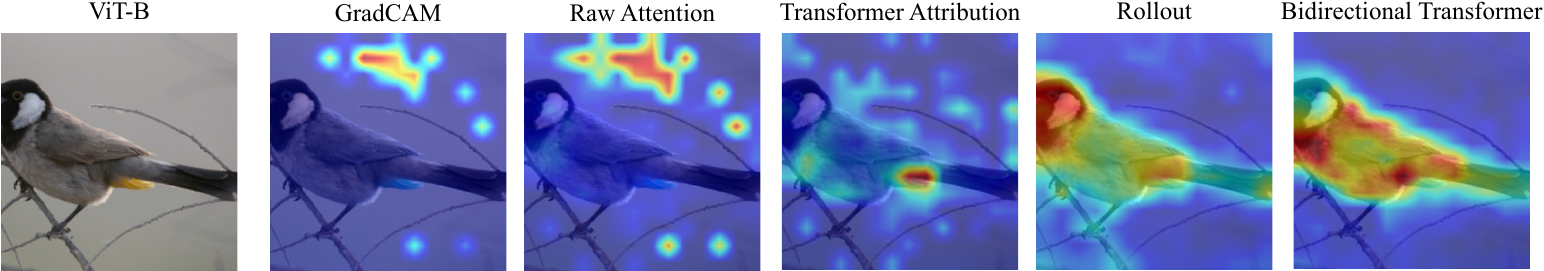}} \
    \subfloat[ViT-B-DINO]{\includegraphics[width=0.48\textwidth]{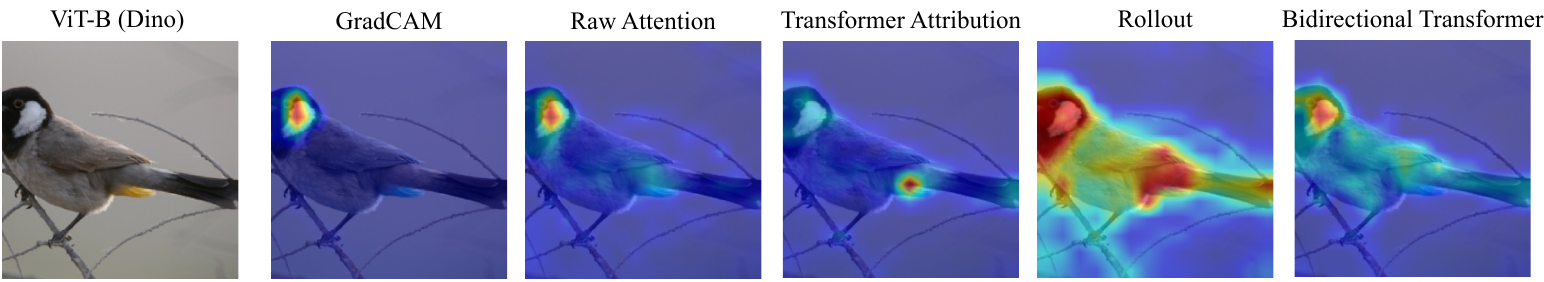}}\\ \vspace{-3mm}
    \subfloat[ViT-B-MAE]{\includegraphics[width=0.48\textwidth]{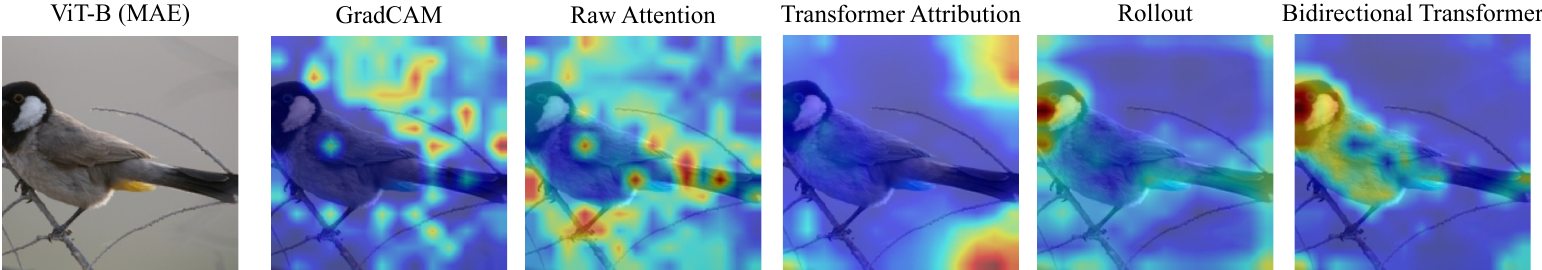}} \
    \subfloat[ViT-B-MOCO]{\includegraphics[width=0.48\textwidth]{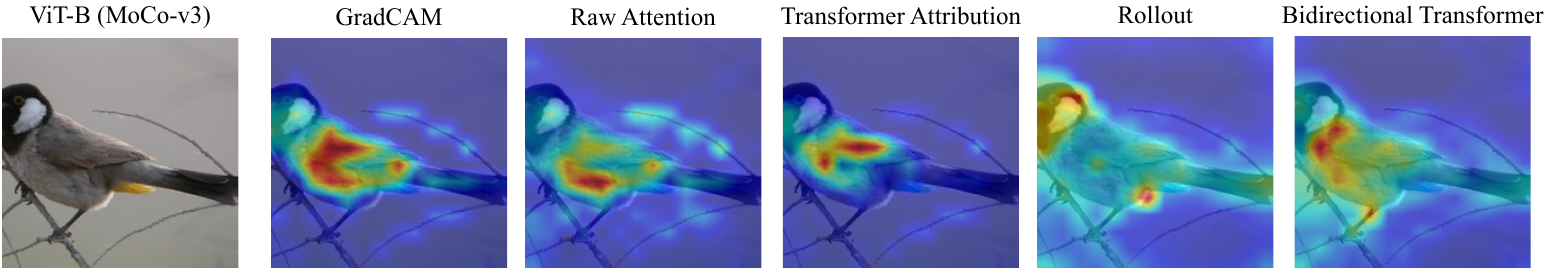}}\\ \vspace{-3mm}
    \subfloat[BERT-based Model]{\includegraphics[width=0.95\textwidth]{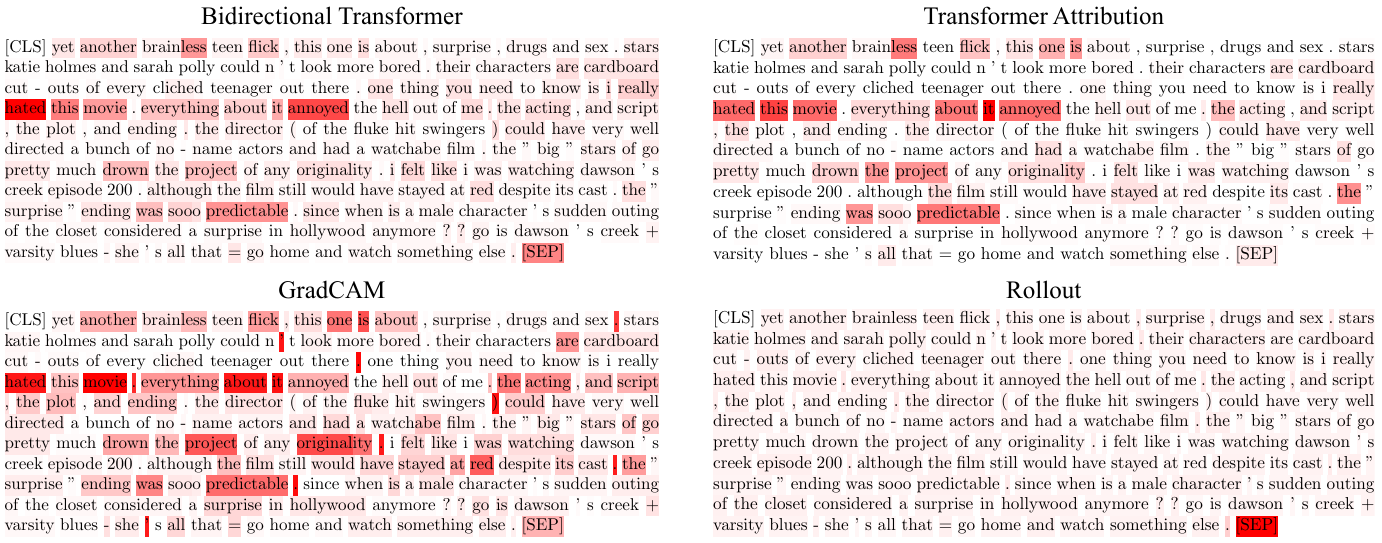}}
    \caption{Visualization of Commonly-used Feature Attribution Methods with Vision and NLP Models: (a)--(d) the ViT-base model and derivatives fine-tuned for birds classification~\cite{wah2011caltech}; (e) a BERT model fine-tuned on IMDb movie reviews~\cite{maas-EtAl:2011:ACL-HLT2011}. }
    \label{fig:feature_attr}
     \vspace{-5mm}
\end{figure*}

\subsubsection{Activation/Attention-based Methods}
Utilizing attention and activation mechanisms, XAI aims to improve the interpretability of these DNN models, providing feature attributions into their decision-making processes. In CNNs, feature attributions could be obtained through activation mappings. Class Activation Mapping (CAM)~\cite{zhou2016learning} employs global average-pooled feature maps to highlight relevant regions for prediction, whereas Grad-CAM~\cite{selvaraju2017grad} utilizes gradients from convolutional layers to create visual explanations for feature attributions, illustrating important image regions affecting model outcomes.

Transformers, based on self-attention mechanisms, adeptly capture sequential dependencies without the constraints of recurrent architectures~\cite{vaswani2017attention}. XAI probes these attention distributions to infer feature significance. Techniques like Attention Rollout and Attention Flow~\cite{abnar2020quantifying} aggregate attention scores across layers, revealing pathways that predominantly inform the model's output. Similarly, Layer-Wise Relevance Propagation (LRP) applied to Transformers~\cite{voita2019analyzing} backtracks the relevance scores from the output to the input tokens, thereby identifying the contribution of features. Following this line of research, more recently, Bidirectional attention flow~\cite{chen2022beyond}, transformer attribution~\cite{hao2021self} and attribution rollout~\cite{xu2023attribution} have been proposed to leverage attention weights inside transformers to estimate the feature attributions for particular prediction. 

For example, Fig.~\ref{fig:feature_attr} visualizes a suite of feature attribution techniques, showcasing their application to various models. The observations underscore the variable nature of feature attribution outcomes depending on the chosen method, underlining the necessities for faithfulness tests to validate fidelity of these methods. Li~\emph{et al.}, however, discovered the consensus of visual explanations across different image classifiers~\cite{li2021cross} and leveraged such consensus as pseudo labels of segmentation tasks~\cite{li2023distilling}.

\subsubsection{Proxy Explainable Models}
\begin{figure}
    \centering
    \subfloat[Global Surrogate]{\includegraphics[width=0.4\textwidth]{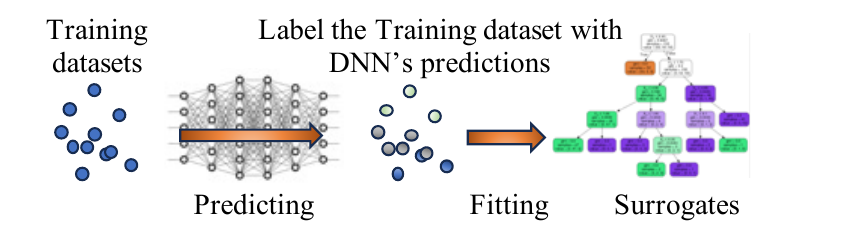}}\\ \vspace{-3mm}
    \subfloat[Local Surrogate]{\includegraphics[width=0.5\textwidth]{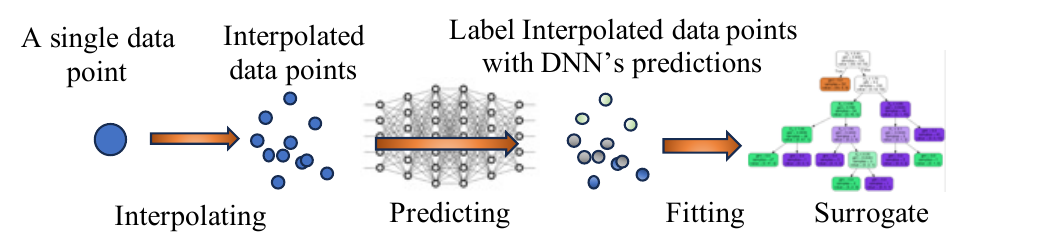}}
    \caption{An Example of Proxy Explainable Models with Global and Local Surrogates for Global and Local Interpretations}
    \label{fig:proxy-explainable}
    \vspace{-5mm}
\end{figure}

Proxy explainable models employ interpretable surrogates to interpret complex DNN models, adhering to the principle of simulating the decision boundaries of the original models with simpler constructs. These surrogate models utilize the predictions from the original DNN as labels to train and elucidate the significance of input features and elucidate their interactions in driving model outputs~\cite{guidotti2018survey}. As shown in Fig.~\ref{fig:proxy-explainable}, two main types of surrogate interpretations can be listed as follows.
\begin{itemize}
    \item \emph{Global Surrogate:} Applied to training or testing datasets, these models yield insights into the overall behavior of the DNN~\cite{frosst2017distilling,di2019surrogate}.
    \item \emph{Local Surrogate:} Targeting a specific input instance, these models aim to decode the rationale behind the model's predictions in the vicinity of that point~\cite{ribeiro2016should,li2023g}.
\end{itemize}
Therefore, surrogate models can embrace either global or local interpretations, sometimes bridging gaps to perturbation-based methodologies~\cite{li2022interpretable}.

Decision trees or ensemble tree methods such as random forests are well-regarded surrogates due to their inherent interpretability, enabling them to establish logical rules linking DNN inputs to outputs, and are typically engaged for global interpretations~\cite{frosst2017distilling,kontschieder2015deep}. These models are proficient in statistically pinpointing feature attributions~\cite{strobl2008conditional,lundberg2018consistent}.

Linear surrogates, despite their presumption of linearity, which contrasts with the non-linearity of deep models, remain valuable, particularly in spotlighting the linear contributions of different features via coefficients~\cite{harrell2001regression}. Notable among the linear approaches are LIME and G-LIME, which distill the decisions made by classifiers or regressors into local linear approximations~\cite{ribeiro2016should,li2023g}. Polynomial models, expanding this toolkit, unravel complex decision patterns by applying methods akin to Deep Taylor Decomposition (DTD), which approximates the model's output contributions by backpropagating relevance with local Taylor expansions, an effective tool for demonstrating the importance of each feature progressively across layers~\cite{montavon2017explaining}. 

In summary, while decision trees and forests are suitable for global surrogates in echoing the non-linear nature of DNNs, techniques like Random Forest may distill DNN behavior to determine broader feature importance~\cite{strobl2008conditional,lundberg2018consistent}, as elaborated in instructional resources~\cite{global-surrogate,local-surrogate}.

\subsection{Reasoning Process as Model Explanation}
To explore the model's internal decision-making pathways, representative methods have been proposed as follows.

\subsubsection{Visualizing Intermediate Representations}
Visualizing intermediate representations in deep learning models is crucial to elucidating how these models process information. Intermediate feature visualization translates complex transformations in hidden layers into an interpretable format, revealing the key patterns that the model focuses on for predictions~\cite{zeiler2014visualizing}.

Network Dissection evaluates the interpretability of deep visual representations by correlating neurons with semantic concepts, identifying the alignment based on intersection over union (IoU) metrics, thus interpreting neuron activations in a human-understandable manner~\cite{bau2017network,zhou2018interpreting}. Deconvolutional Network and related methods interpret layers by mapping features to the pixel space~\cite{zeiler2014visualizing,mahendran2015understanding}. These reconstructions serve to reverse engineer the learned representations. Voita~\emph{et al.} propose the strategy focusing on stimulating neurons to high activation and inspecting input modifications, providing insights into learned patterns by neurons~\cite{voita2019analyzing}. Similarly, DeepVis and DeepDream visually dissect neuron activations to interpret what has been learned~\cite{yosinski2015understanding,couteaux2019towards}. 

Eigen-CAM leverages principal component analysis for class-specific activation maps in CNNs~\cite{muhammad2020eigen}, whereas methods proposed by Quach et al. utilize gradient information to refine these visualizations for better class representation~\cite{quach2023using}. These techniques extend beyond basic heatmap visualizations by highlighting activated image segments, thus pinpointing specific learned image features~\cite{zeiler2014visualizing}.

In contrast to feature attribution methods centered on saliency maps, the discussed techniques predominantly focus on the visualization of intermediate features, such as activation maps, to trace the decision-making process within DNN models~\cite{nguyen2016multifaceted}.

\subsubsection{Visualizing the Logic of Reasoning}
Decision trees and their ensemble counterparts, such as random forests and gradient boosting trees, serve as surrogate models to elucidate the decision-making logic of DNNs. These algorithms construct interpretable proxies that replicate the predictions of a DNN by transforming its complex reasoning into a series of simple, logical decisions~\cite{boz2002extracting,frosst2017distilling,zhangquanshi2019interpreting}. Surrogate models leverage inputs and outputs from the DNN, employing the feature attributions these black-box models provide, thereby translating the neural network into an intelligible set of rules or pathways~\cite{barlow2001case}.

By representing the model's decision logic as a branching structure where each node encodes a rule based on feature values, these tree-based algorithms parse the complex reasoning process of DNN into a comprehensible form, culminating in predictions at the leaf nodes~\cite{quinlan1987generating,de2013decision}. This property gives the models a notably intuitive and interpretable nature, allowing the end-to-end decision process to be traced and visualized with ease~\cite{benard2021interpretable}.

Although proxy explainable models specifically quantify feature contributions to a model's output and thus yield local explanations, tree-based surrogate models create a global approximation to the logic of the original model, reflected through an interpretable set of decision rules~\cite{strobl2008conditional,lundberg2018consistent}. These surrogate models offer a broad mimicry of the DNN's inferential pathways, and though they may not individually attribute feature importance, they encapsulate the logic of a model in a structure amenable to human interpretation. Together, these methodologies enrich our capacity to understand the inner workings of DNNs by revealing the foundation and path of their inferences.

\begin{figure}
    \centering
    \includegraphics[width=0.5\textwidth]{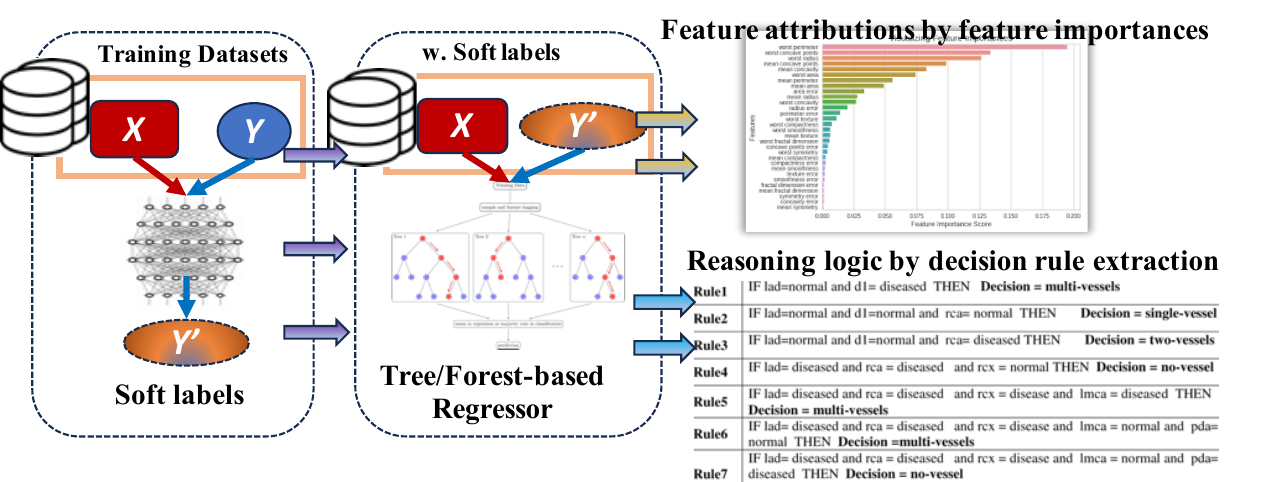}
    \caption{Visualizing feature importance and logic of reasoning with tree/forest-based surrogates}
    \label{fig:enter-label}
    \vspace{-5mm}
\end{figure}

\subsubsection{Counterfactual Examples as Decision Rules}
Counterfactual examples have emerged as an intuitive method for interpreting model decisions in a ``what-if'' analysis~\cite{golfarelli2006designing}, i.e., predicting what would happen if some input to the model changes. These approaches usually frame the decision process into optimization problems with constraints~\cite{chang2018explaining, mothilal2020explaining, goyal2019counterfactual}. These examples aim to identify the least amount of change necessary to the input data to flip a model's prediction, providing clear insights into the model's decision boundaries. FIDO~\cite{chang2018explaining} emphasizes generating counterfactuals that align with fixed feature constraints, while DiCE~\cite{mothilal2020explaining} promotes the creation of diverse counterfactuals, facilitating an understanding of the model across a range of plausible scenarios. 

The incorporation of causal inference perspectives into counterfactual reasoning has further enhanced model interpretability~\cite{moraffah2020causal, xu2020causality}. This approach delves into the causal mechanisms of a model's predictions, identifying which changes in features would lead to different outcomes~\cite{goyal2019counterfactual, laugel2019unjustified}. Adversarial examples are related to counterfactuals in that they also employ optimization techniques to modify inputs, albeit with an aim to challenge a model's robustness~\cite{ignatiev2019relating}. They reveal the model's vulnerabilities and provide insights into the learning process~\cite{geirhos2020shortcut, ilyas2019adversarial}. In addition to generating counterfactuals, some interactive visualization tools that perform ``what-if'' hypothesis testing via factors controlling on data can also provide analyses based on counterfactual reasoning~\cite{wexler2019if}.

Both perturbation-based methods and counterfactual reasoning look at how input modifications influence the model's output. However, counterfactual examples focus on the \emph{``minimal modification''} to change the prediction, offering a causal inference on the functionality of models~\cite{goyal2019counterfactual, laugel2019unjustified}. Through this, one can appreciate the underlying decision logic of DNN models from a causality-driven point of view~\cite{pearl2009causal}.

\subsubsection{Prototypes as Decision Rules} 
 Mining prototypes from training data serves as a method to distill and interpret the decisions of deep learning models by identifying exemplars or representative features~\cite{chen2019looks,guidotti2019black,nauta2021looks,donnelly2022deformable,singh2022think,nauta2023co,wang2023hqprotopnet}. The idea of class-specific exemplars dates back to previous work~\cite{bien2011prototype}, but recent approaches, like ProtoPNet~\cite{chen2019looks}, actively learn prototypes from the last layer of a deep model to provide understandable associations between the model's decisions and the training data. These prototypes offer visual explanations for class predictions, enhancing the interpretability of networks.

Deformable ProtoPNet expands on this by capturing prototypical parts and contemplating pose variations and context, thus enriching both the accuracy and the interpretability of the model~\cite{donnelly2022deformable}.  Nauta~\emph{et al.} proposed a benchmark to evaluate the performance of image classification and interpretability based on ProtoPNet~\cite{nauta2023co}. The introduction of support and trivial prototypes~\cite{wang2023hqprotopnet} further helps to understand the behavior of DNN models at decision boundaries.

For global interpretability, PIP-Net~\cite{nauta2023pip} introduces an aggregation of learned prototypes that transparently illustrates the reasoning process of the model, affording professionals the means to audit the robustness of pattern recognition applications in critical fields such as medical image analysis~\cite{mohammadjafari2021using,carloni2022applicability,nauta2023interpreting}. 

These prototype-based methods primarily target the creation of self-explaining classifiers~\cite{gautam2023looks} rather than explaining existing DNN models. To bridge this gap, Nauta~\emph{et al.}~\cite{nauta2021looks} introduced a technique to explain the prototypes derived from ProtoPNet-like models. Furthermore, the interpretability of these prototypes has recently been explored through human-centered analyses~\cite{kim2022hive,davoodi2023interpretability,huang2023evaluation,behzadi2023protocol}.

\subsubsection{Concept Activation Vectors and Derivatives}
 Concept Activation Vectors (CAVs) provide interpretable dimensions within the activation space of a neural network, representing abstract ``concepts'', ranging from objects to colors~\cite{kim2018interpretability}. A CAV is formalized as a vector orthogonal to a hyperplane, discriminating activations with or without the concept. Building upon CAVs, Testing with Concept Activation Vectors (TCAV) offers a quantitative approach to assess the influence of particular concepts on model predictions~\cite{kim2018interpretability}. The TCAV score, derived from directional derivatives along the CAV, signifies the degree to which a concept is implicated in the model's output. A positive derivative implies a positive TCAV score, with the class-specific score computed as the proportion of instances positively associated with the concept.

 Moreover, recent advancements have seen the introduction of invertible CAVs to interpret vision models with non-negative CAVs~\cite{zhang2021invertible}, and Text2Concept which extends the CAV framework to NLP, allowing the extraction of interpretable vectors from text~\cite{moayeri2023text2concept}. Concept Activation Regions (CARs) further generalize this framework by using a collection of CAVs to define the decision boundaries in DNN models~\cite{crabbe2022concept}.

% ~\cite{selvaraju2017grad,hao2021self,chen2022beyond,xu2023attribution} 

\subsection{Summary and Discussion}
In summary, given the features as input, XAI techniques can interpret the model's decision either from the perspectives of \emph{feature attribution} or \emph{reasoning process}. Here, we map some representative methods of these two directions into the flow of data mining and discuss them as follows.

\subsubsection{Data Acquisition and Collection}
 In the realm of DNN interpretation, diverse data types are crucial for implementing methods such as LIME~\cite{ribeiro2016should}, G-LIME~\cite{li2023g}, Feature Ablation~\cite{merrick2019randomized, fong2019understanding, ramaswamy2020ablation}, and SHAP~\cite{lundberg2017unified}, among others. These methods adeptly tackle tabular, text, and image data to interpret the decision-making processes. For tabular and text data, these interpretability techniques elucidate the contribution of each feature to the prediction outcome, treating each variable as an interpretable feature. In the context of image data, the focus shifts to uncovering the significance of individual or clusters of pixels in model predictions, with G-LIME employing strategies for superpixel clustering to construct the features for attributions.

\subsubsection{Data Preparation and Transformation}
Distinct methodologies for data transformation are vital for interpreting DNNs. LIME and G-LIME generate data perturbations around local instances~\cite{robnik2018perturbation}, while Feature Ablation sets input features to a predetermined baseline. SHAP traverses all combinatorial feature subsets during training. Input baselines for Integrated Gradients~\cite{qi2019visualizing,lundstrom2022rigorous} and multiple noisy instances in SmoothGrad~\cite{smilkov2017smoothgrad} are other approaches to acquire interpretable data. Furthermore, intermediate DNN representations can be obtained using gradient-based attribution~\cite{ancona2018towards}, layer-wise relevance propagation~\cite{montavon2017explaining}, and techniques like Network Dissection~\cite{boz2002extracting,frosst2017distilling,zhangquanshi2019interpreting}, investigating the representations learned by layers. Derivatives of CAM~\cite{selvaraju2017grad,chattopadhay2018grad,wang2020score,jiang2021layercam,muhammad2020eigen} also facilitate the extraction of features from networks.

\subsubsection{Data Modeling and Analyses}
The modeling of data and subsequent analyses vary across interpretability methods. LIME and G-LIME fit explainable models on perturbed data, mainly for local feature importance inference. Decision trees and related nonlinear rule-based models offer comprehensive global interpretations. Feature ablation evaluates the impact of feature omission, while SHAP employs a game-theoretic methodology, quantifying the marginal contribution of each feature~\cite{roth1988shapley}. Integrated Gradients calculate the path integral for feature influence clarification. SmoothGrad averages multiple noisy input gradients for a stable interpretation. In addition, approaches such as counterfactuals or concept activations utilize gradient exploration for DNN decision boundary exposition~\cite{kim2018interpretability}. Deep Taylor Decomposition backtracks neuron outputs to input signals to ascertain feature relevance, fostering an in-depth analysis for model interpretation.

\subsubsection{Results Reporting \& Visualizations}
Visualization strategies are central to reporting in image-based data interpretation. Techniques such as LIME, G-LIME, SHAP, DTD, Integrated Gradients, and SmoothGrad use heatmaps to highlight significant image regions~\cite{samek2016evaluating}. Additionally, G-LIME, SHAP, and Feature Ablation project feature attributions as ranked lists, highlighting the order over precise values. Feature Ablation visualizes ``ablated images'', indicating critical pixel/superpixel configurations. Intermediate representation visualization may involve saliency or attention maps derived from activation maps. Moreover, counterfactuals employ comparative data rows, while concept activation vectors are elucidated via variance plots and correlation graphs. The chosen visualization approach hinges on both the model's complexity and the interpretive method employed, always striving for interpretative clarity.

%\subsubsection{Discussions}
%While the above methods offer ways to understand how each feature contributes to the prediction made by the model or the reasoning process to make the decision from input data, they can be computationally expensive, results may vary, or they might be biased due to the use of certain priors. Furthermore, the use of simple proxy models or pursing in human-readable interpretation could also be potential pitfalls.

\section{Influences: Data Valuation and Anomaly Detection of Training Samples}
Interpreting deep models by measuring the influence of training samples on decision making is essential for understanding and validating the outputs of these models. The process generally involves several techniques that map the correlation between individual training samples and the decisions made by the model~\cite{sogaard2021revisiting,hammoudeh2022training}. In this section, we categorize existing work into three folders as follows.

\subsection{Sample Valuation as Model Explanation}
Sample Contribution-Based Methods form a distinct category of interpretability techniques that aim to interpret deep models by determining the influence of individual training examples on the model's decision making. The fundamental idea of these methods is to measure how much the prediction for a test instance would change if a particular training instance were not included in the training dataset. Although most of these methods are derived from robust statistics~\cite{cook1980characterizations,hampel2011robust,bates2023cross}, we summarize some as follows.

\subsubsection{Gradient-based Methods for Valuation} The advent of influence functions~\cite{cook1980characterizations,hampel2011robust} has become a crucial analytical tool in XAI, gauging the sensitivity of model predictions to marginal changes in training data~\cite{koh2017understanding,feldman2020does}. This technique quantifies the hypothetical exclusion of individual data points, informing on their respective contribution to the final model outcome. Its application spectrum is broad, facilitating the introspection of language models~\cite{grosse2023studying}, metric learners~\cite{liu2022debugging}, and performance refinement with subset training~\cite{wang2020less}. Although invaluable for identifying outliers and influential instances, thus bolstering model robustness, the computational toll of influence functions remains prohibitive for extensive datasets~\cite{hampel2011robust}. Guo~\emph{et al.} proposed FastIF, which can scale-up influence function estimation over large-scale NLP datasets for both explanation and debugging purposes~\cite{guo2021fastif}. Bhatt~\emph{et al.} introduce a fast yet robust conformal prediction method that approximates multiple leave-one-out classifiers through influence function~\cite{bhatt2021fast}.

TracIn~\cite{NEURIPS2020_e6385d39} ushers in a complementary paradigm, harnessing backpropagation to determine the influence of training samples on predictions, a key to diagnosing biases and reinforcing fairness in vision and language models~\cite{NEURIPS2021_13d7dc09, NEURIPS2022_d0702278}. It also helps classifier calibration by flagging anomalous training inputs~\cite{pmlr-v162-lin22h}. Unlike gradient-based techniques, its robustness transcends model constructs, offering insights into non-differentiable ensemble learners~\cite{brophy2023treeinfluence}. However, the efficacy of \textit{TracIn} is contingent on data integrity; skewed or noisy datasets could impair the reliability of its inferences. Alongside, its underlying assumption of loss function smoothness can be an ill-fit for complex neural architectures, calling for caution in extrapolation.

The formidable computational investment associated with gradient and Hessian computations~\cite{bae2022if} requires optimization to feasibly scale these interpretative tools to high-dimensional DNN models.

\begin{figure}
    \centering
    \includegraphics[width=0.45\textwidth]{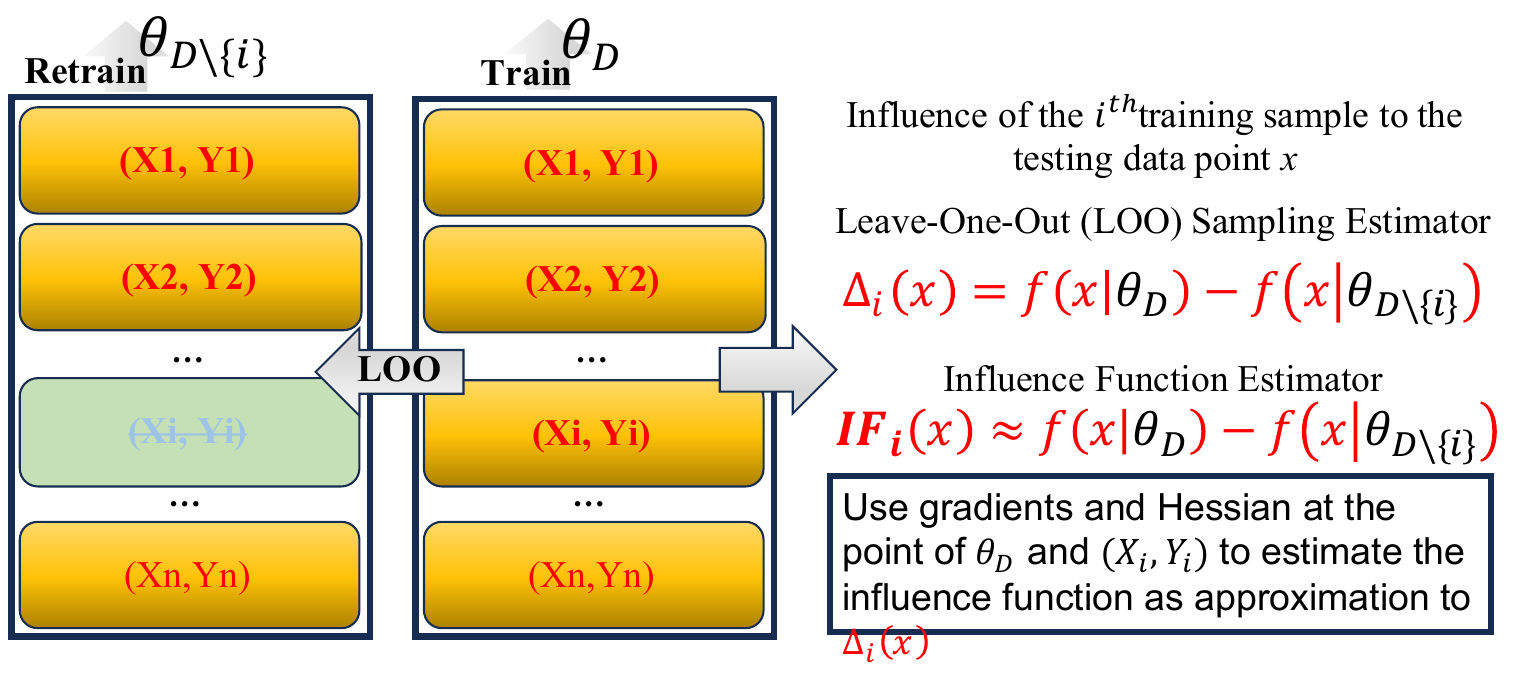}
    \caption{Evaluating the influence of a training sample: Leave-One-Out (LOO) resampling versus influence functions}
    \label{fig:loo-vs-if}
    \vspace{-5mm}
\end{figure}

\subsubsection{Resampling-based Methods for Valuation} 
 Resampling strategies such as Leave-One-Out (LOO) and Jackknife Resampling are crucial in gauging the contribution made by individual data points to predictive models. The LOO approach operates on the principle that the impact of each data point can be evaluated by excluding the data point from the training set and observing the resultant variations in the predictions~\cite{bates2023cross}. Its comprehensive nature allows for the detection of outliers and influential samples that disproportionately affect the model~\cite{black2021leave, ye2023leave}, although at the cost of computational efficiency, particularly with large datasets.

Jackknife Resampling extends this evaluative framework by introducing statistical measures to quantify the influence wielded by each observation when omitted~\cite{efron1992jackknife}. This method exhibits superior performance over traditional influence functions in addressing the complexity of DNN responses~\cite{basu2020influence, alaa2020discriminative}. Compared to LOO, Jackknife estimates provide a more tractable approximation of the sampling distribution while avoiding the need to retrain the model with each resample~\cite{alaa2020frequentist, sattigeri2022fair}. However, the method assumes linearity and may fail in the face of non-linear data structures or outliers~\cite{friedl2002jackknife}.

Fig.~\ref{fig:loo-vs-if} illustrates a simple comparison between influence function and LOO resampling, where both strategies intend to evaluate the influence of a sample to decision making---the resampling-based method needs to re-train a model with the sample excluded and test the model, while the influence function-based approach can directly measure the influence by computing gradients or Hessians. The trade-off between these two types of approaches rely on the scale of datasets and computational complexity of differentiation.

%Collectively, these resampling-based methods serve as indispensable tools in the realm of XAI, providing critical insights into data point significance at the expense of heightened computational demand, and must negotiate the complexity and nonlinearity of large-scale training datasets.  

\subsubsection{Game theoretical Methods for Valuation} 
Shapley values provide a robust framework for estimating the contribution of individual training samples to a model's predictions, extending the existing applications to feature attribution~\cite{rozemberczki2022shapley}. Data Shapley operationalizes this by linking training samples directly to model outputs, offering a measure for the influence of data points on predictive outcomes~\cite{ghorbani2019data}. Beta Shapley further refines this process, presenting a noise-reduced data valuation that accelerates calculations while maintaining critical statistical attributes~\cite{kwon2022beta}.
 
In a parallel vein, the use of Banzhaf values has recently gained momentum in the field of data valuation~\cite{wang2023data}. These values employ the Banzhaf power index to evaluate the influence of a sample on model predictions, highlighting the advantage of maximizing the safety margin between Shapley values and LOO errors. In recent developments, DataInf emerges as a non-gradient, non-resampling game-theoretical methodology, particularly well suited to understanding sample influence in fine-tuned large language models~\cite{kwon2023datainf}. This is complemented by work exploring the overarching concept of global sample importance~\cite{tsai2023sample}, providing a more macroscopic perspective on the significance of the data points' influence.
 
Model-free approaches such as LAVA and CS-Shapley hold the potential for alternative, potentially more practical influence estimation, relying on class-wise Wasserstein distances or class-specific Shapley values, respectively~\cite{just2022lava, schoch2022cs}. When these methods substitute actual labels with those predicted by the model, they can effectively estimate the influence of training samples on decision-making processes, thus improving the analytical toolkit available for data valuation and model interpretation.

\subsection{Sample Anomalies as Model Explanation}
In addition to valuate the contribution of training samples to specific decision making, there also exists a line of research focusing on mining the samples with anomalies from the training dataset subject to the predictions of trained models. Here, we categorize these works by their application purposes as follows.

\subsubsection{Hard Sample Mining} 
Hard Example Mining (HEM) is an algorithmic strategy aimed at identifying the most difficult-to-learn subsets training samples, enhancing the learning process in machine learning models \cite{zhou2023samples}. Stemming from statistical learning concepts \cite{hastie2009boosting}, HEM employs mechanisms such as those in AdaBoost, which uses residual prediction errors to modify sample weights, thereby improving generalization capabilities \cite{schapire2013explaining}. Boosting methods \cite{friedman2001greedy,chen2016xgboost,ke2017lightgbm}, through this tactically calibrated weight adjustment, effectively address diverse statistical learning challenges.

\begin{figure}
    \centering
    \subfloat[Forgetting Events]{\includegraphics[width=0.25\textwidth]{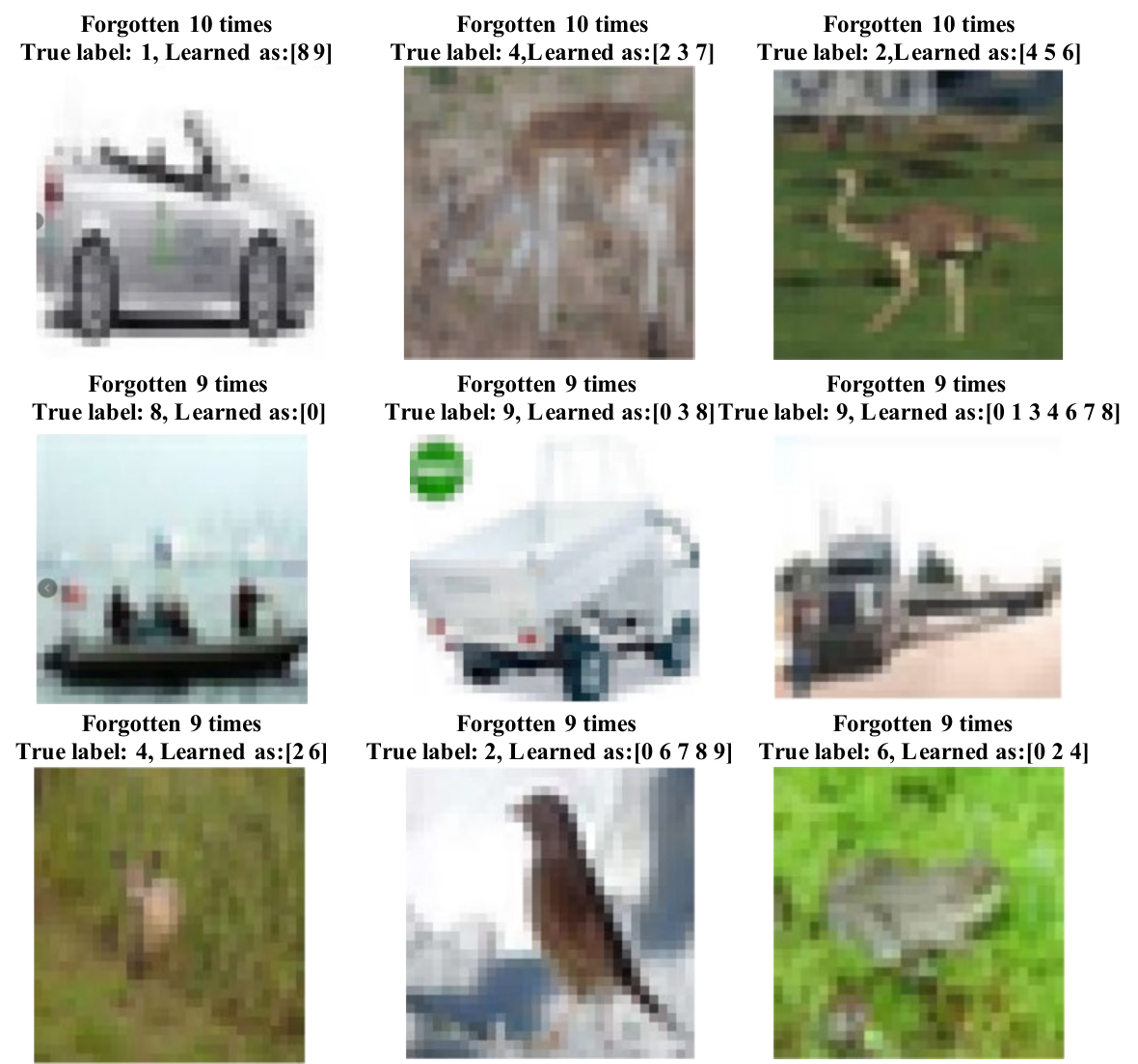}} \ 
    \subfloat[``Never Forgotten'']{\includegraphics[width=0.23\textwidth]{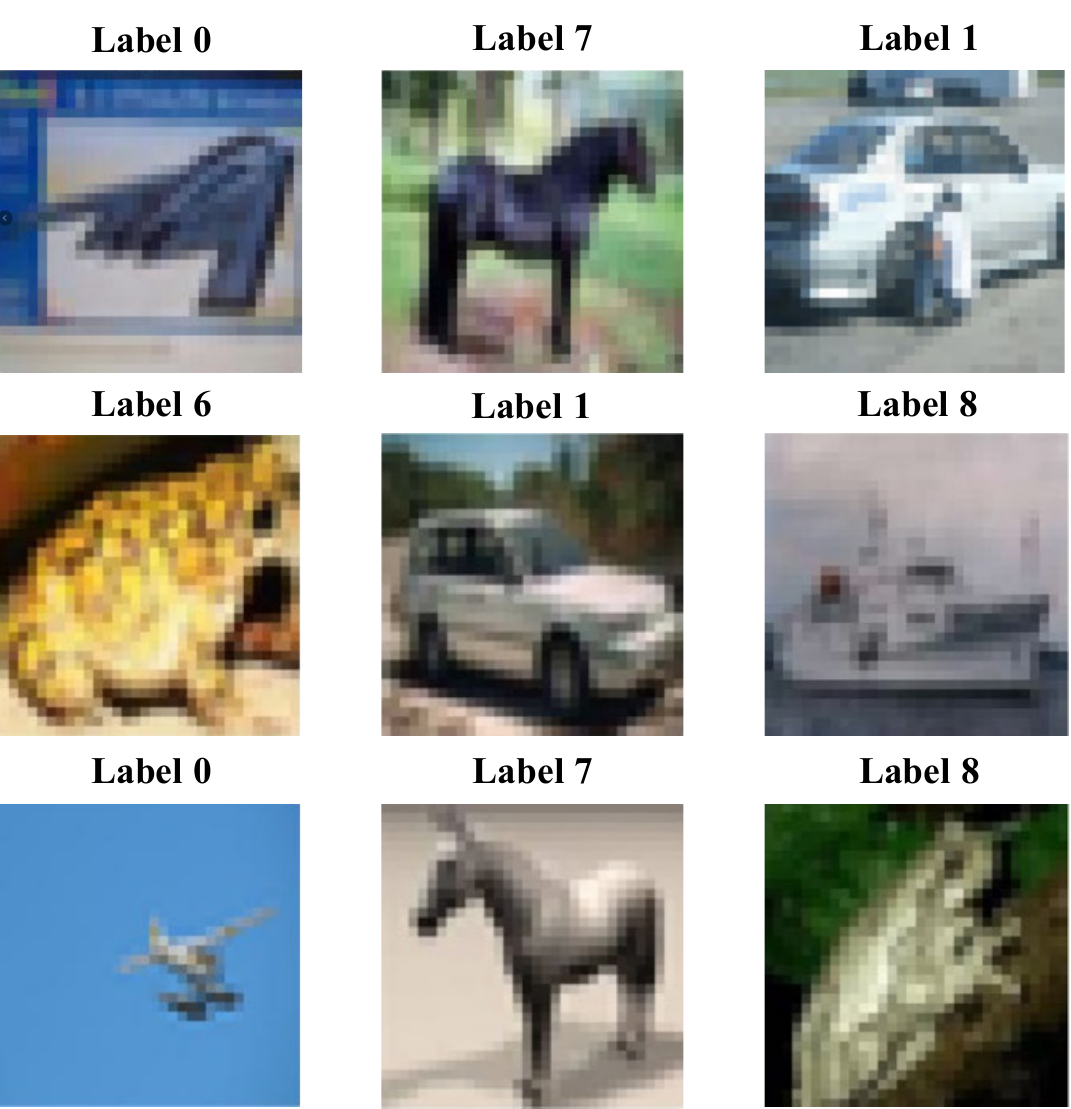}}%\\ \vspace{-3mm}
    \caption{Samples that suffer forgetting events and samples that were never forgotten during the training of CIFAR-10.}
    \label{fig:forgetting-events}
    \vspace{-5mm}
\end{figure}

In the context of Support Vector Machines (SVMs), ``hard samples'' correspond to support vectors situated nearest to the defining hyperplane, marking them as pivotal training instances \cite{cortes1995support,ma2014support}. The principle of identifying challenging samples transcends the traditional applications and is now a staple in computer vision research \cite{canevet2016large,shrivastava2016training,dong2017class,smirnov2018hard,suh2019stochastic,tang2019uldor,sheng2020mining,zhao2020deep}. Implementations of HEM in these fields have resolved data bottlenecks and addressed class imbalance through algorithms like Monte Carlo Tree Search (MCTS) for difficult sample recognition \cite{canevet2016large} and online HEM enhancing object detection \cite{shrivastava2016training}. Moreover, innovations such as auxiliary embeddings have augmented HEM efficacy in image dataset applications \cite{smirnov2018hard}, while detection of forgetting events~\cite{toneva2018empirical} can also help to discover hard samples (or mislabeled ones), where the forgetting event is defined as the moment that an individual training sample transitions from being classified correctly to incorrectly over the course of learning. Fig.~\ref{fig:forgetting-events} illustrates the examples of forgetting events and the samples that were never forgotten during the training procedure of CIFAR-10. Through tracing the forgetting events, one can either identify the hard samples from the early stage of training~\cite{paul2021deep} or measure the difficulty of every training sample ~\cite{baldock2021deep}.

Deep learning for computer vision, dealing with class imbalances, has also integrated HEM into specialized loss functions like Focal loss \cite{lin2017focal} and triplet loss \cite{chechik2010large}, which implicitly refine training through a focus on challenging samples. Extending past its roots in computer vision, HEM is now being leveraged across assorted domains such as active learning, graph learning, and information retrieval, embracing the challenge of elucidating and concentrating on complex data elements in their respective fields \cite{liu2023hard,formal2022distillation,zhang2022m}. These recent forays into varied domains represent a broadening scope and utility of hard sample mining in enriching the learning paradigms across the data science spectrum.

\subsubsection{Mislabel Detection}
The task of detecting, eliminating, or correcting mislabeled samples in training datasets is critical in machine learning, with methodologies broadly classified into three categories~\cite{brodley1999identifying,valizadegan2007kernel,sun2007identifying,xiang2019interactive,pleiss2020identifying,pulastya2021assessing,feng2021phases,jia2023learning}.
 
Firstly, visual and interactive analytics are prominent and utilized to identify mislabeled data points by detecting outliers in low-dimensional representations~\cite{xiang2019interactive,zhang2020dealing}. Techniques for detection of out-of-distribution (OOD) samples also fall into this category, employing visual methods to identify data points that deviate from expected patterns~\cite{chen2020oodanalyzer}. Interactive visualizations have proven to be effective in domains such as health data, where exploratory analyses can lead to improved label accuracy~\cite{ponnoprat2022explainable}.

Secondly, model-based analysis blends algorithmic detection with human oversight~\cite{valizadegan2007kernel,huang2019o2u,xie2021partial}. Kernel methods, for instance, consider the conditional distributions of data and labels, pinpointing mislabels by detecting distribution anomalies~\cite{valizadegan2007kernel}. However, Kernel methods face scalability issues with large-scale deep learning datasets. In response, methods like O2u-Net address mislabel detection in large image datasets~\cite{huang2019o2u}, and approaches for multi-label learning tasks focus on noisy label identification to refine learning~\cite{xie2021partial}. 

\begin{figure}
    \centering
    \subfloat[Training Dynamics and Hard/Mislabeled Samples]{\includegraphics[width=0.5\textwidth]{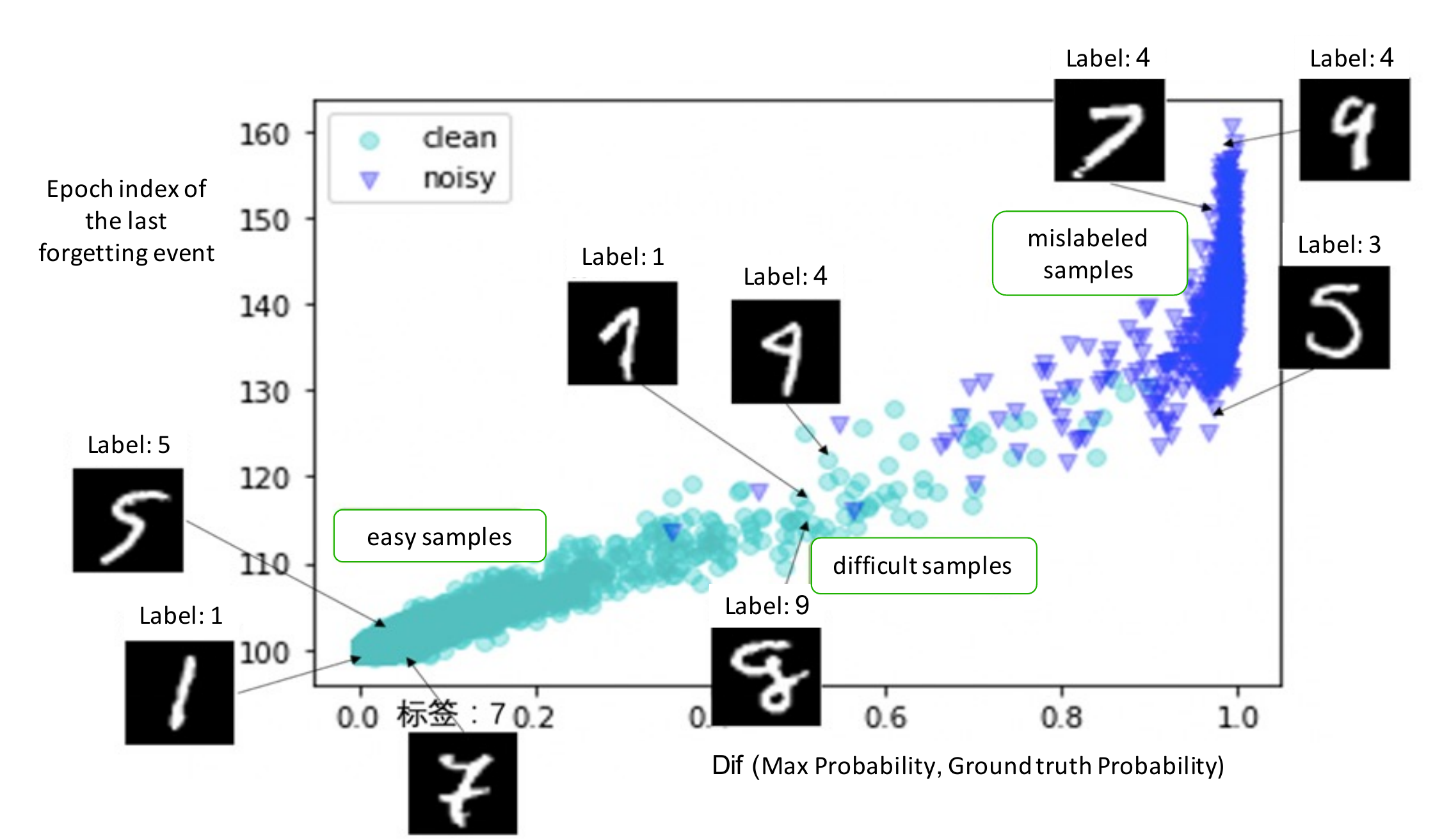}}\\
    \subfloat[Mislabel detection by training dynamics classification~\cite{jia2023learning}]{\includegraphics[width=0.48\textwidth]{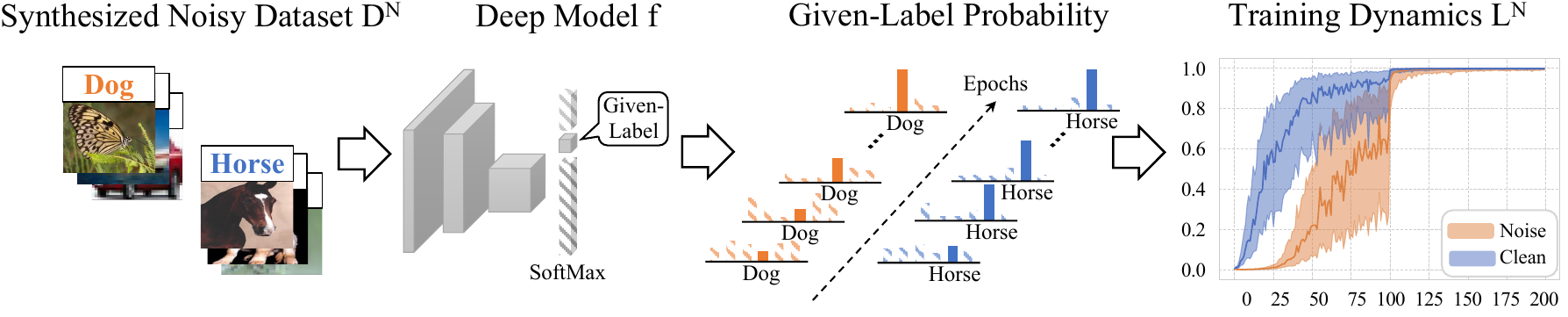}}
    \caption{Training Dynamics and Mislabeled Sample Detection}
    \label{fig:train_dyn}
    \vspace{-5mm}
\end{figure}

Lastly, recent inquiries examine the impact of mislabeled samples on the training processes of deep learning models. Understanding these effects is crucial for developing algorithms capable of mitigating the influence of such samples on model performance~\cite{feng2021phases}. Baldock~\emph{et al.} proposed to map the training samples with the training dynamics (an example on the MINST dataset shown in Fig.~\ref{fig:train_dyn}~(a)) and diagnose the dataset by finding mislabel samples~\cite{baldock2021deep}, where the training dynamics was defined as the evolution of probability difference between the ground truth and the prediction labels of every training sample for every training epoch. Furthermore, as shown in Fig.~\ref{fig:train_dyn}~(b), Jia~\emph{et al.} proposed a four-stage approach: (1) synthesizing mislabeled (noisy) data, (2) training a model with noisy data, (3) collecting data of training dynamics, and (4) classifying the time series of training dynamics to detect mislabeled samples~\cite{jia2023learning}. Theoretical aspects on the training dynamics of noisy-labeled data were investigated in~\cite{xiong2023doubly}.

Together, these methodologies form a multifaceted approach to the challenge of mislabel detection, each contributing unique insights and tools to ensure data integrity in machine learning pipelines~\cite{guan2013survey,gupta2019dealing,song2022learning}.

\subsubsection{Dataset Debugging}
The integrity of training datasets is fundamental to the efficacy of machine learning models, particularly for real-world applications~\cite{jain2020overview, gupta2021data}. Therefore, dataset debugging strategies are crucial, focusing on correcting issues such as missing values~\cite{biessmann2019datawig, cinquini2023handling}, detecting outliers~\cite{sejr2021explainable, yepmo2022anomaly}, and correcting noisy data to improve the performance of the deep neural network~\cite{afzal2021data, gralinski2019geval, zhang2018training}. In handling missing values, Cinquini \emph{et al.} present a post-hoc explainability approach~\cite{cinquini2023handling}, while the work of Sejr \emph{et al.} on outlier detection frames the challenge as an unsupervised classification problem, offering explainable insights~\cite{sejr2021explainable}. Traditional kernel-based methods and statistical tests such as KDE~\cite{cao2005goodness} and the Kolmogorov-Smirnov test~\cite{glazer2012learning} address data distribution issues but can falter with unstructured data types. 

Recent developments in XAI have created new methods to counteract these shortcomings. Techniques for OOD sample detection using post-hoc explanations have been advanced by Martinez \emph{et al.}~\cite{martinez2022can} and TAPUDD~\cite{dua2023task}, with Holmberg \emph{et al.} utilizing visual concepts for this purpose~\cite{holmberg2023exploring} and De \emph{et al.} applying weighted mutual information for OOD identification~\cite{de2023weighted}. Zhang \emph{et al.} proposed an approach to first identify the correctly labeled items and then minimally adjust the remaining labels to produce accurate predictions~\cite{zhang2018training}. To facilitate practical application, MLInspect by Grafberger \emph{et al.}~\cite{grafberger2022data} is noteworthy, providing a means to diagnose and correct technical biases in ML pipelines. Furthermore, PalM was introduced by Krishnan \emph{et al.} through exploring interactive debugging tools~\cite{krishnan2017palm}.

\subsection{Summary and Discussions}
In summary, XAI techniques offer the ability to quantify the impact of training samples on model decisions, viewed through either the \emph{data valuation} or \emph{anomaly detection} lens. Here we map representative methodologies belonging to these categories within the data mining process and provide subsequent discussion.

\subsubsection{Data Acquisition and Collection}
Within the domains of \emph{data valuation} and \emph{anomaly detection}, XAI methodologies proffer a comprehensive framework to discover the influence of training data on predictive modeling, exhibiting adaptability across various data types such as tabular, text, and image~\cite{koh2017understanding, basu2020influence}. \emph{Sample valuation-based} explainers such as influence functions, TracIn, LOO, Jackknife, and Shapley Valuation are engineered to accommodate a wide array of data, with TracIn showing substantial applicability in image and text scenarios~\cite{koh2017understanding, NEURIPS2020_e6385d39, martinez2002visual, basu2020influence, ghorbani2019data}. Conversely, \emph{sample anomaly-based} explainers are typically optimized for specific data modalities, with approaches like O2u-Net, TAPUDD, and Smirnov et al. centering on image data~\cite{huang2019o2u, dua2023task, smirnov2018hard}, while Jia et al. offer a versatile methodology that is effective across data formats~\cite{jia2023learning}, and PaLM focuses on tabular datasets~\cite{krishnan2017palm}.

\subsubsection{Data Preparation and Transformation} 
Various \emph{sample anomaly-based} explainers apply various preprocessing techniques to prepare data for analysis. Smirnov et al. produce auxiliary embeddings for images~\cite{smirnov2018hard}, while O2u-Net utilizes an oscillatory training method to capture training loss~\cite{huang2019o2u}. Jia et al. shuffle labels and track per-epoch training loss~\cite{jia2023learning}, TAPUDD employs heatmap extraction from DNN models~\cite{dua2023task}, and PaLM partitions datasets using decision tree nodes~\cite{krishnan2017palm}. These preparatory steps enrich XAI applications, enhancing the understanding of the influences of the training data on the model outputs.

\subsubsection{Data Modeling and Analysis}
In data modeling and analysis, XAI is used to unravel the impact of training data on predictive models. Sample valuation-based explainers such as influence functions~\cite{koh2017understanding} and TracIn~\cite{NEURIPS2020_e6385d39} utilize gradients in their assessments, while methods like LOO~\cite{martinez2002visual} measure changes in model predictions after sample exclusion. On the anomaly detection front, Smirnov et al. pair challenging samples during training~\cite{smirnov2018hard}, O2u-Net ranks samples by loss curves highlighting potential mislabels~\cite{huang2019o2u}, and Jia et al. employ a time series classifier to spot mislabeling~\cite{jia2023learning}. TAPUDD detects outlying examples via heatmap clustering~\cite{dua2023task}, and PaLM provides global explanations through decision tree partitions~\cite{krishnan2017palm}, facilitating a greater understanding of the influence from training data on model decisions.

\subsubsection{Results Reporting and Visualization} 
Result presentation varies across XAI tools. Influence functions compute the sway of individual training samples, facilitating pointed insights~\cite{koh2017understanding}, whereas TracIn quantifies both positive and negative training influences~\cite{NEURIPS2020_e6385d39}. Techniques like LOO display influence by showing prediction shifts~\cite{martinez2002visual}, while Jackknife and Shapley Valuation assign values signifying influence magnitude~\cite{basu2020influence, ghorbani2019data}. On the contrary, the tools~\cite{smirnov2018hard, huang2019o2u} intensify visual clarity in output representation, with PaLM showcasing iterative model responses~\cite{krishnan2017palm}. These methods ensure critical training data elements are visually scrutinized for their impact on models, helping to systematic exploration of the influence landscape.

%\subsubsection{Discussions}
%Above methods, while powerful for estimating the influence of data to complex models, commonly face the drawback of requiring significant computational resources, particularly with large datasets. Additionally, their performance may be hampered if the quality of the underlying data is poor, with some methods also potentially assuming linearity in complex data sets which may not always hold true. The selection of an appropriate XAI approach, therefore, demands a careful consideration of these limitations in the context of the task at hand.  

\section{Insights: Patterns and Knowledge Discovery from Data}
XAI algorithms facilitate the extraction of human-readable insights, in part by identifying and explaining patterns, correlations, and anomalies in complex multi-dimensional or multi-modal data. Two groups of efforts have been done: one concerning societal values and the other focused on the advancement of scientific discovery.

\subsection{Societal Values as Model Explanation}
XAI here % refers to methods and techniques developed in the field of AI research that make the output of AI or machine learning models understandable by humans. It 
aims to improve the interpretability and trustworthiness of algorithms and models making decision for social choices, which in turn can improve social fairness, ethics, accountability, and transparency.

\subsubsection{Algorithmic Fairness}
XAI is crucial to unlock and improve biases within machine learning models, allowing equity in automated decision-making as evidenced by the advancement of Counterfactual Fairness approaches \cite{madras2018learning}. The transparency provided by XAI is instrumental in clarifying deep neural network logic, laying the foundation for accountable and equitable AI applications \cite{zhou2023explain}. Furthermore, the dissection of fairness narratives, especially within microlending and educational systems, underscores the need to take into account diversified stakeholder perspectives \cite{smith2023many, gardner2023cross}. Benbouzid et al. \cite{benbouzid2023fairness} and Shrestha et al. \cite{shrestha2023help} enunciate the importance of optimizing AI to embrace scientific fairness and effective collective decision making.

The practical application of XAI to fairness is explored by Castelnovo \emph{et al.}, who examine the limitations of conventional group fairness mitigation techniques in dynamic financial environments and offer a novel strategy fusing human behavioral insights for ongoing model refinement \cite{castelnovo2021towards}. Stanley \emph{et al.} highlight the role of XAI in uncovering subgroup bias in medical imaging, where disparities in XAI outputs provide clues to underlying biases \cite{stanley2022fairness}. Lastly, Alikhademi \emph{et al.} present a tripartite framework to evaluate XAI tools on bias detection and mitigation, revealing that while XAI can explain model behavior, its capabilities in bias detection are often wanting \cite{alikhademi2021can}.

\subsubsection{Digital Ethics}
The confluence of ethics and XAI epitomizes a growing area of inquiry. McDermid~\emph{et al.}~\cite{mcdermid2021artificial} foreground the influence of data-driven decision making on ethical practices within societal infrastructures, a position supported by Rakova and Dobbe \cite{rakova2023algorithms}, who advocate for an environmental justice lens in algorithmic auditing. This ethos integrates XAI within a broader systemic framework, advocating for ethical engagement to achieve equitable resolutions.

Juxtaposed ethical considerations between AI professionals and the public surface in Jakesch~\emph{et al.}~\cite{jakesch2022different}, which highlight the discrepancy in responsible AI value prioritization. The authors posit that comprehensible XAI systems are instrumental in bridging this divide, thus cultivating trust and alignment with social norms. Mccradden et al. \cite{mccradden2023s} contribute JustEFAB guidelines to advance clinically transparent machine learning models, advocating for a lifecycle perspective in bias examination and rectification framed within ethical and justice theories. This precept mirrors the imperative process of XAI to elucidate AI-driven decision-making to endorse ethical clinical applications. Counterfactual reasoning in XAI undergoes an investigation from Kasirzadeh et al. \cite{kasirzadeh2021use}, highlighting potential risks in misapplying immutable social categorizations, thus encouraging cautious development in accordance with the principles of fairness and transparency.

AI benchmarking, viewed in a position paper \cite{blili2023making}, as analogous to human intelligence assessments, illuminates the subjective nature of AI success metrics. The authors argue for AI benchmarks as pivotal constituents of the XAI ambit to uphold ethical AI growth through recognition and mitigation of inherent biases. Hawkins and Mittelstadt \cite{hawkins2023ethical} examine the ethical quandaries of AI data enrichment, advocating the crystallization of research ethics standards to guide this area, thus underscoring the crucial role of XAI in navigating ethical complexities.

\subsubsection{Systems Accountability}
 AI accountability is paramount for equitable, legal, and socially aligned decisions, fostering human involvement in critical decision-making and averting unwarranted reliance on technology. Kacianka et al. shed light on AI accountability through Structural Causal Models (SCMs), providing a methodology that reveals the layers of accountability within the domain of autonomous vehicles, enhancing transparency and reliability of decision-making \cite{kacianka2021designing}.
 
 Hutchinson et al. propose an accountable framework for the lifecycle of machine learning datasets, ensuring traceable and accountable decision-making, thus contributing significantly to AI audit practices and development \cite{hutchinson2021towards}. Kroll et al. explore traceability in XAI, linking it to global AI policies and prescribing a set of requirements for accountable AI systems. Through identifying technological shortcomings, they chart a path towards improved traceability and thus improved accountability in AI practices \cite{kroll2021outlining}.
 
 Lima et al. address accountability in autonomous XAI applications, cautioning against the misuse of explanations as liability shields. The paper urges robust regulatory frameworks to maintain ethical standards in AI, with an emphasis on sensitive fields such as healthcare and employment \cite{lima2022conflict}. Donia et al. explore normative frameworks within XAI, articulating various perspectives including verification, representation, social license, fiduciary duty, and legal compliance. This dissection of approaches helps ensure that AI technologies satisfy ethical, social and legal standards, leading to superior AI governance and design in accordance with XAI responsibilities and societal expectations \cite{donia2022normative}.

\subsubsection{Decision Transparency}
The advancement of XAI in various sectors addresses the need for transparency and trust in AI systems. In healthcare, ``Healthsheet'' contributes to this by providing detailed dataset documentation, specifically subject to healthcare data peculiarities, revealing biases and promoting data source transparency \cite{rostamzadeh2022healthsheet}. 

Schmude et al. emphasize the need for understandable algorithmic explanations to establish trust and fairness in AI applications \cite{schmude2023impact}, while TILT goes further by offering a transparency language that meets and exceeds GDPR requirements, thus supporting data protection clarity \cite{grunewald2021tilt}. In education, XAI has been instrumental in clarifying the AI-driven NYC school admission process, facilitating parental engagement and contributing to accountability and fairness \cite{marian2023algorithmic}.This case study demonstrates the ability of XAI to promote transparency and verification within public policy, discovering mechanisms behind student placements through crowd-sourced parental involvement, which bolsters accountability, fairness and decision transparency. 

Lastly, Chile initiates to apply algorithmic transparency standards in public administration reflects a commitment to making automated government decisions open and accountable, fostering citizen trust \cite{lapostol2023algorithmic}. This project systematically evaluates the use of automated decision systems by government against a transparency framework, underscoring the need for these systems to be intelligible and accessible by the public.

\subsection{Scientific Explorations as Model Explanation}
Rather than focusing on societal issues of AI, XAI provides practical techniques and tools that hold significant potential for scientific discovery and research \cite{zednik2022scientific,krenn2022scientific,li2021kepler}. Here, we categorize these works into three groups by their approaches of leveraging XAI techniques.

\subsubsection{Pattern Recognition} 
Scientific datasets are instrumental in modern research, serving as the basis for uncovering novel patterns through data mining techniques \cite{kamath2001mining, karpatne2017theory}. Deep learning excels at detecting complex patterns within these datasets but often functions opaquely, making the extraction of information by researchers challenging. Addressing this opacity, XAI interventions provide elucidation on AI decision-making processes \cite{zednik2022scientific}, increasing the utility of deep learning in areas such as drug discovery by ensuring model transparency and intelligibility \cite{jimenez2021coloring}. 

Incorporating XAI techniques such as concept whitening (CW) into graph neural network models not only improves trustworthiness but also improves interpretability and performance in the drug discovery field \cite{proietti2023explainable}. By tailoring CW for spatial convolutional GNNs and related graph network frameworks, the self-interpretability of the models is fortified, subsequently improving drug efficacy predictions. The field of computational chemistry benefits from the role of XAI in mitigating deep learning model obscurities, promoting an understanding of molecular structure-property relationships such as solubility and permeability \cite{wellawatte2023perspective}. Utilization of counterfactual explanations and attribution methods like SHAP and gradient-based techniques elucidates the complex associations learned by deep learning models. Such XAI applications cement confidence in predictions and inform molecular adjustments to engineer desired characteristics, substantially advancing the capacity of deep learning to identify and predict chemical phenomena.

Machine learning methods, when paired with XAI, have also played a part in reinvigorating classical scientific principles, exemplified by the re-conceptualization of Newton's law of gravitation \cite{lemos2023rediscovering}. Analysis of extensive celestial datasets via graph neural networks has heightened prediction accuracy for planetary motion, enhancing confidence in the derived insights.  

\subsubsection{Interdisciplinary Collaboration} 
 Interdisciplinary collaboration is paramount for translational research, blending a diverse range of expertise to foster cross-disciplinary innovation and speeding up the translation of lab findings into practical applications \cite{weston2010faculty,guerrero2017interdisciplinary}. XAI paves the way for such synergies among various domains, including computer science, physical and chemical sciences, and medicine, leading to cross-disciplinary breakthroughs \cite{boden2008mind}. Advanced AI techniques contribute to these interdisciplinary explorations, developing integrative models that propel scientific advancement \cite{wang2023interpretable, zednik2022scientific}.

 In particular, the role of XAI in biophysics and computational science is highlighted by the achievements of the AlphaFold project in protein folding \cite{jumper2021highly}. Employing the XSMILES visualization tool, XAI has significantly improved molecular property prediction, elucidating complex molecular patterns through interactive visualization, and helping experts understand model behaviors and comparative model evaluation \cite{heberle2023xsmiles}.
 In medical imaging, XAI acts as a conduit between deep learning algorithms and clinicians, revealing disease indicators in MRI scans, which are crucial for precise diagnostics yet can be undetectable to the naked eye \cite{el2021multilayer, essemlali2020understanding, li2022interpretable}. Such transparent AI systems are vital in healthcare for accurate and early disease detection and diagnosis. 
 
 Furthermore, XAI is essential for collaborative translational medicine, merging deep learning with clinical practice. It is utilized in designing models for lung cancer detection from CT scans, uncovering both known and novel diagnostic patterns \cite{ardila2019end}. AI systems like IBM Watson for Oncology also demonstrate how AI can help clinicians craft personalized treatment plans, prioritizing transparency for healthcare professionals and patients \cite{basu2011developing}. Borys et al. highlighted the need for nonvisual XAI methods to overcome the constraints of current visual explanations in medical imaging, aiming for better alignment with the interpretability required by medical practitioners \cite{borys2023explainable}. 
 AI frameworks are similarly applied to decipher tinnitus patient data, translating complex patterns into clear guidelines for clinicians, thus improving individualized treatments and overall patient care \cite{tarnowska2021explainable}. XAI is instrumental in converting scientific progress into applied clinical practice, marrying data-driven insights with actionable clinical decisions.

\subsubsection{Uncovering Mechanisms} 
In the pursuit of advancing science across disciplines, XAI has become crucial to deciphering the complexities of natural systems and enhancing the depth and precision of scientific investigations \cite{burkart2021survey,carleo2019machine}. The utility of machine learning algorithms in disentangling complex system dynamics is particularly evident in quantum physics, where AI aids in understanding quantum entanglement and state dynamics, with practical applications in the creation of quantum devices \cite{gross2017quantum}.

An example on effectiveness of XAI is the use of a Restricted Boltzmann Machine (RBM) by the University of Bristol, which improved the detection of quantum entanglement, allowing researchers to recognize patterns learned by RBM across various quantum systems \cite{melko2019restricted}. In genomics, AI has played a role in elucidating gene regulation complexities, thereby deepening the comprehension of gene interactions and regulatory mechanisms \cite{keyl2023single,park2020global}.

Sahin \emph{et al.} \cite{sahin2021xai} have stressed the importance of XAI in uncovering hidden relationships within photovoltaic materials, confirming that feature engineering combined with XAI can optimize predictions for organic solar cell efficiency. This promotes interdisciplinary collaboration in materials science. Likewise, Anguita \emph{et al.} \cite{anguita2020explainable} use XAI to identify gene expression patterns related to obesity by employing rule-based algorithms that offer insights into gene-gene interactions derived from longitudinal observations.

In the field of geosciences, XAI helps in approximating the water content in minerals such as clinopyroxene, an essential factor for understanding Earth water budget \cite{li2023explainable, rozemberczki2022shapley}. The adoption of interpretative tools, like Shapley values, clarifies the role of various elements in hydrogen diffusion within these minerals. Consequently, XAI demystifies AI models, helping scientists validate or refute hypotheses and foster novel insights \cite{li2023explainable}.

As scientific inquiry often involves the integration of various methodologies to investigate natural phenomena, XAI stands out as a promising avenue for such interdisciplinary examination \cite{zednik2022scientific}. Despite the novelty and untapped potential of XAI in scientific research, recognizing the opportunities presented by AI, and particularly XAI, is essential for the future of scientific exploration \cite{lavin2021simulation}.

\subsection{Summary and Discussions}
XAI for advancing societal values and scientific discovery, when viewed through the lens of data mining stages, integrates the four steps as follows.

\subsubsection{Data Collection and Acquisition}
Studies reviewed leverage a wide array of complex multi-dimensional or multi-modal datasets for XAI-enabled analysis. For societal applications, data types include demographic, financial, medical imaging, behavioral, and educational datasets, helping to address issues such as algorithmic fairness~\cite{madras2018learning,zhou2023explain,castelnovo2021towards}, digital ethics~\cite{mcdermid2021artificial}, and system accountability~\cite{kacianka2021designing}. In the realm of scientific discovery, data types range from genomic sequences~\cite{keyl2023single} to quantum system properties~\cite{gross2017quantum}, passing through molecular structures for drug discovery~\cite{wellawatte2023perspective} and clinical imaging data for healthcare applications~\cite{el2021multilayer}.

\subsubsection{Data Preparation and Transformation}
Prior to XAI-enabled analysis, the data must undergo rigorous pre-processing. In societal contexts, pre-processing steps include standardization to mitigate bias, ensure fairness and transparency~\cite{stanley2022fairness}, and transform inconsistent or missing financial data for dynamic settings~\cite{castelnovo2021towards}. For scientific exploration, studies can employ normalization, feature extraction and selection techniques to hone the datasets relevant to the domain, such as chemical attributes for drug discovery~\cite{proietti2023explainable} or physiological and environmental factors for medical diagnostics~\cite{ardila2019end}.

\subsubsection{Data Modeling and Analysis}
In terms of modeling and analysis, XAI is deployed through a variety of techniques aligned with the target task. For societal fairness, Counterfactual Fairness approaches and deep neural network transparency techniques are utilized to clarify logic and mitigate biases~\cite{madras2018learning}, while rule-based algorithms and Structural Causal Models (SCMs) are utilized for systems accountability in decision-making processes~\cite{kacianka2021designing}. In scientific discovery, pattern recognition through XAI involves concept whitening with graph neural networks for drug efficacy prediction~\cite{proietti2023explainable}, while Restricted Boltzmann Machines (RBMs) help unravel quantum entanglement~\cite{melko2019restricted}. These methods model and analyze the data to extract interpretable patterns, fostering trust and deepening understanding.

\subsubsection{Results Reporting and Visualization}
The reporting and visualization of XAI results vary widely based on context. In societal applications, visualizing fairness in healthcare datasets is achieved through subgroup analysis~\cite{stanley2022fairness}, and public transparency is promoted through comprehensive documentation like ``Healthsheet''~\cite{rostamzadeh2022healthsheet}. Scientific findings are similarly conveyed through intuitive visualization tools, such as the XSMILES tool for molecular property prediction~\cite{heberle2023xsmiles}, or through non-visual methods requiring further development, as in medical imaging~\cite{borys2023explainable}. The clarity and interpretability of models’ decision-making processes are enhanced across domains, from aiding medical professionals in diagnosis to assisting policy-makers in evaluating algorithms for public administration~\cite{marian2023algorithmic}. Overall, the deployment of XAI in both societal and scientific arenas emphasizes interactive and accessible visualizations optimized for the respective audience to enhance understanding and application of AI insights.

\section{Limitations and Future Directions}
 In summary, we identify key technical limitations in the domain of explainable AI (XAI) methods as follows.

 \textbf{Data Quality}: The efficacy of XAI methods depends on the quality of the data. Poor data quality can lead to less effective explanations. To address this, a focus on robust data preparation and transformation is critical to improve data quality and ensure more reliable results~\cite{jain2020overview}.

\textbf{Algorithmic Complexity}: With the increasing complexity of AI models, explaining their decisions without oversimplification poses a significant challenge. Advanced XAI technologies must evolve to efficiently unpack complex models, especially with large datasets, without compromising interpretability~\cite{basu2020influence,chuang2023efficient}.

\textbf{Method Evaluation and Selection}: Current evaluation frameworks may not fully capture the wide spectrum of XAI methods, leading to a possible misjudgment of their utility. Developing expansive and thorough evaluation frameworks is essential to inform better method selection, ensuring high-quality explanations across different models and scenarios~\cite{mohseni2021multidisciplinary,agarwal2022openxai,li2023mathcal}.

\textbf{XAI at LLM Scale}: The scalability and complexity of Large Language Models (LLMs) such as GPT-4 mean that current XAI approaches may not adequately address biases or errors in training data of these models. The research is leaning towards creating scalable XAI strategies to provide clear explanations for the decisions made by such expansive models~\cite{grosse2023studying,zhao2023explainability}.

\section{Conclusions}
This work presents a structured review of the role of Explainable AI (XAI) through the lens of data mining, addressing three vital thematic areas:
\begin{itemize}
    \item \textbf{Interpreting Model Behaviors}: This review underscores the imperative of uncovering the decision-making processes of Deep Neural Networks (DNNs) from the perspective of feature attributions and reasoning logic, aiming to increase the transparency and trust in AI systems.
    \item \textbf{Evaluating Data Influences}: This review focuses on how individual data samples shape the model's decision and generalization performance, spotlighting the significant contributors to learning and detecting any data anomalies that might lead to skewed outcomes.
    \item \textbf{Distilling Actionable Insights}: Beyond providing explanations, the review seeks to discover new insights that align with societal values and promote scientific innovation, steering the knowledge from XAI techniques towards practical applications.

\end{itemize}
In conclusion, the study conducts a thorough analysis of XAI methods for the above three purposes, highlighting current capabilities, practical uses, and detecting areas that need improvement. The analysis sets the stage for further research that strives to integrate XAI more deeply into data mining practices and to cultivate a more transparent, accountable, and user-centric AI landscape.

\bibliographystyle{IEEEtran}
\bibliography{main}
\clearpage
\appendices

\section{Study Designs and the Taxonomy System}
To conduct a systematic survey of ``explainable AI'' from data mining perspectives, we follow the Preferred Reporting Items for Systematic Reviews and Meta-Analyses (PRISMA) approach~\cite{moher2009preferred}. The study is designed as follows.

\subsection{Related Surveys}
As an emerging area of research, several surveys have already been conducted for XAI. In this section, we review the existing surveys in this area. 

Known for their ``black-box'' nature, the deciphering of deep models poses a significant academic interest as underscored in the work of Doshi-Velez and Kim \emph{et al.} which emphasizes the need for evaluation techniques for XAI methods~\cite{doshi2017towards}. The survey by Carvalho \emph{et al.}~\cite{carvalho2019machine} gives a comprehensive look at the range of interpretability methods and metrics that exist within machine learning. They delve into both model-agnostic and model-specific interpretability techniques, while also highlighting the various metrics utilized in gauging interpretability. The applicability of XAI to specific domains has also been explored, such as in the field of drug discovery, as highlighted in the work of Preuer \emph{et al.}~\cite{preuer2019interpretable}. This research showcases the potency of XAI methods in improving the understanding of deep learning models, thus crucial within scientific discoveries. Tjoa and Guan~\cite{tjoa2020survey} provide another significant domain-specific perspective on XAI, detailing the use of XAI techniques in the context of medical image analysis. Their work is of significant relevance, allowing for the translation of these models' sophisticated diagnostics to be more digestible for healthcare professionals. 

Additionally, the comprehension of complex statistical models such as random forests has been addressed in the study by Haddouchi and Berrado~\cite{haddouchi2019survey}. They overview a myriad of interpretability methods, tailoring them for direct application within random forest models. Exploring the practical paradigms of XAI, Marcinkevi{\v{c}}s and Vogt~\cite{marcinkevivcs2023interpretable} and Notovich \emph{et al.}~\cite{notovich2023explainable} present a variety of XAI methods within machine learning, while cementing these with application-specific examples and providing a taxonomy of different XAI techniques, respectively. The comprehensive nature of these works bridges the gap between theory and real-world applications. Meanwhile, Hammoudeh and Lowd~\cite{hammoudeh2022training} provide a distinct perspective, shifting the attention from model interpretability to the role of training data in models. They detail the impacts and various methods for determining the importance of training data contributions. 

%Despite the variety of approaches to XAI and its application across various domains, common challenges remain. The balance between interpretability and model complexity, the selection of suitable evaluation metrics, and the unique interpretation requirements of different applications demand intensive research moving forward. This is crucial in improving the interpretability, transparency, and trustworthiness of machine learning models.  

\subsection{Research Methods}
Despite the variety of approaches to XAI and its application across various domains, common challenges remain. The balance between interpretability and model complexity, the selection of suitable evaluation metrics, and the unique interpretation requirements of different applications demand intensive research moving forward. In this case, our study may be interested in two questions as follows 
 \begin{itemize}
     \item \textbf{Q1.} What are the current problems, methodologies, strengths/weaknesses, outcomes/limitations and future directions of explainable AI techniques? 
     \item \textbf{Q2.} Whether and how they can be formulated into a typical data mining process, consisting of \emph{data acquisition \& collection}, \emph{data preparation \& transformation}, \emph{data modeling \& analyses} and \emph{results reporting \& visualization}?
 \end{itemize}

\subsubsection{Study Selection and Appraisal}
Given the research questions in mind, we move to the next step that defines the criteria for including or excluding studies. We start by identifying relevant studies in databases such as PubMed, IEEE Xplore, Google Scholar, Web of Science, Scopus, EMBASE, and ACM Digital Library. Use a well-composed combination of keywords, for instance: ``Explainable AI'', ``Interpretable AI'', ``Transparent AI'', ``Interpretable Deep Learning'', ``Explainable Deep Learning'', ``Transparent Deep Learning'', ``Explainable Data Science'', ``Survey'', ``Review'', etc. Additionally, perform hand-searching of reference lists of the identified articles and other topic-related reviews.

After carefully gathering a detailed list of publications, we methodically remove any duplicates. A preliminary assessment based on titles and abstracts allows us to exclude articles lacking relevance to the field of Explainable AI within a Data Mining context. Our selection criteria aim to ensure that all chosen papers inherently enhance our research objectives. Specifically, the papers included should: 
\begin{itemize} 
\item Propose XAI algorithms with meaningful involvement in data processing, modeling, and analysis, or 
\item Investigate data-driven methodologies for interpreting the behaviors of DNN models and AI systems, or 
\item Evaluate and augment datasets by contemplating models' decision-making explanations, or 
\item Utilize XAI-focused techniques for dataset analysis geared towards knowledge discovery. 
\item Lastly and crucially, the selected papers must present unique insights, propose innovative techniques, or improve understanding within the domain. 
\end{itemize} 
We then proceed to remove any paper that does not meet these rigorous criteria. The ultimate collection of papers is decided by mutual agreement within the author team. This approach is designed to incorporate various perspectives, thereby restricting potential bias.

\subsubsection{Study Analysis, Grouping and Reporting}
Subsequently, we capture meta-data from each study, comprising elements including authors, publication year, study objectives, methodologies used, core findings, and conclusions drawn. In addition, we assess the quality of the shortlisted papers, with a concentration on their methodology, results, and ensuing discussions. 

We then meaningfully categorize notable findings from these articles, based on recurring themes such as utilized applications, methodologies, successful results, encountered obstacles, among others. We proceed to discuss the major findings, their significance, and their contributions specifically to explainable AI from a data mining vantage point. Lastly, we address limitations observed within the studies and propose directions for future research to promote the advancement of the field.  

%\subsection{Brief on Results and the Taxonomy}

\begin{table*}[]
\centering
\caption{Comparison among Feature Attribution-based Explainers}
\scriptsize
\setlist[itemize]{leftmargin=1.5mm}
%\resizebox{1.0\textwidth}{!}{
\begin{tabular}{|>{\RaggedRight}m{1.5cm}|>{\RaggedRight}m{1.8cm}|>{\RaggedRight}m{3cm}|>{\RaggedRight}m{3.5cm}|>{\RaggedRight}m{3.5cm}|>{\RaggedRight}m{2.2cm}|} \hline
\textbf{Algorithm} & \textbf{Data} & \textbf{Transform.}& \textbf{Model \& Analysis}& \textbf{Results \& Visualiz.}& \textbf{Pros \& Cons}\\ \hline
LIME~\cite{ribeiro2016should} & 
\begin{itemize}
    \item Tabular, 
    \item Texts, 
    \item Images and etc.
\end{itemize}
 & 
 \begin{itemize}
     \item Perturb inputs from local instances;
     \item Collect responses from the model.
 \end{itemize}
  & 
  \begin{itemize}
      \item Fit proxy explainable models (e.g., linear regression via LASSO) using perturbed inputs/responses.
  \end{itemize}
  & 
  \begin{itemize}
      \item Coefficient of proxy models as feature importance scores;
      \item Heatmaps for visualization of image data.
  \end{itemize}
 & 
 \begin{itemize}
     \item Model-agnostic and local explanations;
     \item Computationally expensive, results may vary.
 \end{itemize}
\\ \hline
G-LIME~\cite{li2023g}& 
\begin{itemize}
    \item Primarily for images through superpixel clustering;
    \item Tabular and etc.
\end{itemize}
& 
\begin{itemize}
    \item Cluster superpixels as explainable features;
    \item Follow LIME~\cite{ribeiro2016should} to obtain perturbed inputs and responses
\end{itemize}
& 
\begin{itemize}
    \item Global interpretation as the prior;
    \item Fit global prior-induced ElasticNet as proxy using perturbed inputs/responses.
\end{itemize}
 & 
\begin{itemize}
    \item Use coefficient paths of ElasticNet as importance ranking of features;
    \item Heatmaps for visualization of image data.
\end{itemize} 
& 
\begin{itemize}
    \item Lower complexity for image data due to the use of superpixel clustering;
    \item Bias induced due to the use of global prior.
\end{itemize}
\\ \hline
Feature Ablation~\cite{merrick2019randomized,fong2019understanding,ramaswamy2020ablation}& 
\begin{itemize}
    \item Tabular, 
    \item Texts, 
    \item Images and etc.
\end{itemize}
& 
\begin{itemize}
    \item Replace input features of local instances with a given baseline;
    \item Train a model and collect responses from the model.
\end{itemize}
& 
\begin{itemize}
    \item Measure the change in the model's responses;
    \item Estimate the attribution of every feature.
\end{itemize}
& 
\begin{itemize}
\item  Feature importance for local instances.
\end{itemize}
& 
\begin{itemize}
    \item Computationally intensive, ignorance to feature interactions.
\end{itemize}
\\ \hline
SHAP~\cite{lundberg2017unified}& 
\begin{itemize}
    \item Tabular, 
    \item Texts, 
    \item Images and etc.
\end{itemize}
& 
\begin{itemize}
    \item Train the model using ``all possible'' feature combinations;
    \item Collect responses from the model.
\end{itemize}
& 
\begin{itemize}
    \item Model the competition among features in contribution to responses;
    \item Compute the marginal contribution of every feature.
\end{itemize}
 & 
 \begin{itemize}
    \item Histogram of feature importance for local instances;
    \item Heatmaps for visualization of image data.
\end{itemize}
& 
 \begin{itemize}
    \item Sound game-theoretically;
    \item Computationally intensive due to combinations.
\end{itemize}
\\ \hline
Integrated Gradients~\cite{qi2019visualizing,lundstrom2022rigorous}
& 
\begin{itemize}
    \item Primarily images 
\end{itemize}
& 
\begin{itemize}
    \item Prepare a baseline (usually a image of all zeros) for the input image.
\end{itemize}
& 
\begin{itemize}
    \item Compute the path integral from the baseline to the input image following the direction of gradients.
\end{itemize}
&
\begin{itemize}
    \item Gradients as the contribution of each pixel to the model's prediction, visualized as heatmaps.
\end{itemize}
&  
\begin{itemize}
\item Applicable to many differentiable models;
\item Computationally intensive with high resolutions.
\end{itemize}

\\ \hline
SmoothGrad~\cite{smilkov2017smoothgrad}
& 
\begin{itemize}
\item Primarily images
\end{itemize}
& 
\begin{itemize}
\item Add random noises to the input images;
\item Generate multiple noisy versions of input.
\end{itemize}
& 
\begin{itemize}
\item Compute gradient for every noisy input image;
\item Averages gradients over multiple noisy versions of input.
\end{itemize}
& 
\begin{itemize}
    \item Averaged gradients as the contribution of each pixel to the model's prediction, visualized as heatmaps.
\end{itemize}& 
\begin{itemize}
\item Reduces the volatility of saliency maps, provides more accurate interpretation;
\item May blur significant feature distinctions.
\end{itemize}
\\ \hline
DTD~\cite{montavon2017explaining} & 
\begin{itemize}
\item Primarily images
\end{itemize}
& 
\begin{itemize}
\item Compute the pixel-wise Taylor decomposition of the full neural network as function.
\end{itemize}
&
\begin{itemize}
\item Distributes the output of each neuron back to its input signals;
\item Traceback to the pixel-wise input.
\end{itemize}
& 
\begin{itemize}
\item Provides measure of importance for each input feature.
\end{itemize}
& 
\begin{itemize}
\item Characterize the nonlinear behaviors of the network;
\item Complexity increases with model depth.
\end{itemize}

\\ \hline
\end{tabular}
%}
\label{tab:feature_attr}
\end{table*}

\begin{table*}[]
\centering
\caption{Comparison among Reasoning Process-based Explainers}
\scriptsize
\setlist[itemize]{leftmargin=1.5mm}
\begin{tabular}{|>{\RaggedRight}m{1.5cm}|>{\RaggedRight}m{1.5cm}|>{\RaggedRight}m{3cm}|>{\RaggedRight}m{3.5cm}|>{\RaggedRight}m{3.5cm}|>{\RaggedRight}m{2.2cm}|} \hline
\textbf{Algorithm} & \textbf{Data} & \textbf{Transform.}& \textbf{Model \& Analysis}& \textbf{Results \& Visualiz.}& \textbf{Pros \& Cons}\\ \hline
\hline
Network Dissection~\cite{bau2017network} &
\begin{itemize}
    \item Images
\end{itemize} 
& 
\begin{itemize}
    \item Label and localize semantic concepts in every image;
    \item Align the activation of neurons with semantic concepts.
\end{itemize} 

& 
\begin{itemize}
    \item Estimate IoU between activation maps and semantic concept labels; 
    \item Semantic concept match with high IoU considered as representation.
\end{itemize}
& 
\begin{itemize}
    \item Use IoU of semantic concept match;
    \item Utilize activation maps on input images for visualization.
\end{itemize} 
& 
\begin{itemize}
    \item High interpretability by checking alignment with semantic concepts;
    \item Threshold specification needed for IoU
\end{itemize} 

\\
\hline
Deconvolutional Network~\cite{zeiler2014visualizing} 
& 
\begin{itemize}
    \item Images
\end{itemize}  
& 
\begin{itemize}
    \item Prepare and rescale images for reconstruction purposes.
\end{itemize} 
 & 
 \begin{itemize}
    \item Map and visualize features back to pixel space;
    \item Utilize reverse mapping to pixel space for image reconstruction.
\end{itemize} 
  & 
 \begin{itemize}
    \item Use reconstructed images to visualize the key patterns learned by CNN.
\end{itemize}   &
\begin{itemize}
    \item Enables visualization of activated segments; 
    \item Need to build an additional decoder network for interpretation.   
\end{itemize} 
\\
\hline
Tree-based surrogates for logic~\cite{frosst2017distilling} 
& 
\begin{itemize}
    \item Primarily for tabular data
    \item Images and others
\end{itemize} 
& 
\begin{itemize}
    \item Convert features into categorical types;
    \item Collect responses from the DNN model using input datasets.
\end{itemize} 
& 
\begin{itemize}
    \item Fit the model using decision trees and forests;
    \item Extract decision rules from learned trees and forests.
\end{itemize} 
& 
\begin{itemize}
    \item Visualize the branches of decision trees as decision logic;
    \item Plot decision rules as decision logic.
\end{itemize} 
& 
\begin{itemize}
    \item Enable clear visualization and comprehension on logic;
    \item Don't work well when data is hard to categorization.
\end{itemize} 
\\
\hline
FIDO and DiCE~\cite{chang2018explaining,mothilal2020explaining}
& 
\begin{itemize}
    \item Tabular, 
    \item Texts, 
    \item Images and etc.
\end{itemize}
& 
\begin{itemize}
\item Valuate loss functions with varying input of the model,
\item Reconstruct the landscape of loss function subject to the input of model.
\end{itemize}
& 
\begin{itemize}
    \item Use optimization technique to modify the input data;
    \item Obtain minimal alterations on input data to change the model output.
\end{itemize}
 & 
\begin{itemize}
%    \item Changes in input data along with altered model outputs;
    \item FIDO focuses on compliant alterations with observed features;
    \item DiCE provides a diverse set of counterfactual explanations.
\end{itemize} 
& 
\begin{itemize}
    \item Provide causal inference on how model behavior changes with varied inputs;
    \item Can be complex to calculate minimal modifications.
\end{itemize}
\\ \hline
CAV and TCAV~\cite{kim2018interpretability} 
& 
\begin{itemize}
    \item Tabular, 
    \item Texts, 
    \item Images and etc.
\end{itemize}
& 
\begin{itemize}
    \item Identify the Concept of interest;
    \item Match the concept of interest with a vector (concept activation vector) in a particular layer
\end{itemize}
& 
\begin{itemize}
    \item Use normal to a hyperplane to separate instances based on concept presence;
    \item Use the distances to the concept activation vectors as the contribution of specific concepts.
\end{itemize}
& 
\begin{itemize}
    \item Visualize the contribution of specific concepts to a prediction.
\end{itemize}
& 
\begin{itemize}
    \item Quantify concept contribution to model decision; 
    \item Relies on clear concept definition.
\end{itemize}
\\
\hline
\end{tabular}
\label{tab:reason_proc}
\end{table*}

\begin{table*}[] 
\centering 
\caption{Comparison among Sample Valuation-based Explainers} 
\scriptsize 
\setlist[itemize]{leftmargin=1.5mm}
\begin{tabular}{|>{\RaggedRight}m{1.5cm}|>{\RaggedRight}m{1.8cm}|>{\RaggedRight}m{3cm}|>{\RaggedRight}m{3.5cm}|>{\RaggedRight}m{3.5cm}|>{\RaggedRight}m{2.2cm}|} \hline
\textbf{Algorithm} & \textbf{Data.} & \textbf{Transform.} & \textbf{Model \& Analysis} & \textbf{Results \& Visualize} & \textbf{Pros \& Cons} \\ \hline 
Influence Functions~\cite{koh2017understanding} & 
\begin{itemize}
    \item Tabular, 
    \item Texts, 
    \item Images and etc.
\end{itemize}
& \begin{itemize} 
\item Calculate gradients and Hessian matrices of loss functions. 
\end{itemize} 
& 
\begin{itemize} 
\item Use influence function estimators based on gradients and Hessians.
\end{itemize} 
& 
\begin{itemize} 
\item Compute the influence of each training sample to a testing sample;
\end{itemize} & 
\begin{itemize}
\item Robust to outliers;
\item Computationally expensive due to the use of gradients and Hessians.
\end{itemize}
\\ 
\hline 
 TracIn~\cite{NEURIPS2020_e6385d39,brophy2023treeinfluence} 
 & 
 \begin{itemize}
    \item Primarily for images and text data
\end{itemize} 
& 
\begin{itemize}
    \item Calculate gradients via backpropagation~\cite{NEURIPS2020_e6385d39} or boosting~\cite{brophy2023treeinfluence}
\end{itemize} 
& 
\begin{itemize}
    \item Use gradients to correlate every training samples to the model's outputs
\end{itemize}
& 
\begin{itemize}
    \item Compute influence of every training sample to a testing sample
    \item Identify beneficial or harmful influence.
\end{itemize}
&
\begin{itemize}
    \item Surface biases and help in fair audits;
    \item Performance affected by quality/diversity of data.
\end{itemize}
\\ 
\hline
Leave-One-Out (LOO)~\cite{martinez2002visual,black2021leave} & 
\begin{itemize}
    \item Tabular, 
    \item Texts, 
    \item Images and etc.
\end{itemize}
& 
\begin{itemize} 
\item Perform the iterative process of LOO and re-train a model per iteration;
\item Obtain prediction results per iteration. 
\end{itemize} 
& 
\begin{itemize} 
\item Measure and collect the (change of) prediction results using every re-trained model.
\end{itemize} 
& 
\begin{itemize} 
\item Use the change of prediction results as the influence of every individual training sample to the decision-making.
\end{itemize} & 
\begin{itemize} 
\item Simple and holistic approach, 
\item Time-consuming for LOO re-training on large datasets.
\end{itemize} 
\\ 
\hline 
Jackknife~\cite{basu2020influence,alaa2020discriminative} & 
\begin{itemize}
    \item Tabular, 
    \item Texts, 
    \item Images and etc.
\end{itemize}
& 
\begin{itemize}
    \item Like LOO, separates observation for validation;
    \item Rest of the data configured for training.
\end{itemize}
& 
\begin{itemize}
    \item Model statistical variability of every training sample to estimate the significance of influence.
\end{itemize}
 & 
\begin{itemize}
    \item Jackknife influence values for every training sample;
    \item Indicate importance of training samples.
\end{itemize}
& 
\begin{itemize}
    \item Does not require retraining; 
    \item Not robust to outliers, assumption of linearity.
\end{itemize}
\\ 
\hline 
Shapley Valuation~\cite{ghorbani2019data,kwon2022beta} & 
\begin{itemize}
    \item Tabular, 
    \item Texts, 
    \item Images and etc.
\end{itemize}
& 
\begin{itemize}
    \item Like LOO, separates observation for validation;
    \item Rest of the data configured for training.
\end{itemize}
& 
\begin{itemize}
    \item Model statistical variability of every training sample to estimate the significance of influence.
\end{itemize}
 & 
\begin{itemize}
    \item Jackknife influence values for every training sample;
    \item Indicate importance of training samples.
\end{itemize}
& 
\begin{itemize}
    \item Does not require retraining; 
    \item Not robust to outliers, assumption of linearity.
\end{itemize}
\\ 
\hline
\end{tabular} 
\label{tab:valuation}
\end{table*}

\begin{table*}[]
\caption{Comparison among sample anomaly-based Explainers} 
\scriptsize 
\setlist[itemize]{leftmargin=1.5mm}
\begin{tabular}{|>{\RaggedRight}m{1.5cm}|>{\RaggedRight}m{1.8cm}|>{\RaggedRight}m{3.8cm}|>{\RaggedRight}m{3.2cm}|>{\RaggedRight}m{2.8cm}|>{\RaggedRight}m{2.5cm}|} \hline
\textbf{Algorithm} & \textbf{Data.} & \textbf{Transform.} & \textbf{Model \& Analysis} & \textbf{Results \& Visualize} & \textbf{Pros \& Cons} \\ \hline 
Smirnov et al.~\cite{smirnov2018hard}&
\begin{itemize}
    \item Primarily images
\end{itemize} 
&
\begin{itemize}
    \item Prepare every image with an auxiliary embedding (a low-dimensional vector);
    \item Auxiliary embeddings of two ``hard positive'' examples are far from each other, while auxiliary embeddings of two ``hard negative'' examples are close in cosine-similarity metric.
\end{itemize} 
&
\begin{itemize}
    \item Follow the mini-batch generation of Doppelganger mining~\cite{smirnov2017doppelganger} alike for HEM;
    \item For every sample selected in the batch, pair ``hard positive''/``hard negative'' examples accordingly.
\end{itemize}
&
\begin{itemize}
    \item Batches of samples with ``hard positive'' or ``hard negative'' examples inside,
    \item Every ``hard positive'' or ``hard negative'' example is paired with a randomly selected sample in the batch.
\end{itemize}

&
\begin{itemize}
    \item High extensibility with multiple auxiliary embedding strategies;
    \item High computational complexity for Doppelganger sampling.
\end{itemize}
\\ \hline 

O2u-Net~\cite{huang2019o2u} &
\begin{itemize}
    \item Primarily images
\end{itemize} 
& 
\begin{itemize}
    \item Adjust hyper-parameters (e.g., learning rate) during training to make the status of model transfer from over-fitting to under-fitting cyclically;
    \item Record and collect the losses of each sample during iterations.
\end{itemize} & 
\begin{itemize}
    \item Rank training samples by their training loss curves and mislabel samples are usually with larger training losses.
\end{itemize} 
& 
\begin{itemize}
    \item Ranking list of noisy samples (top samples are mislabeled while bottom ones are clean).
\end{itemize} & 
\begin{itemize}
    \item Easy to implement;
    \item Depend on the prior of mislabel rates.
\end{itemize}\\
\hline
Jia et al.~\cite{jia2023learning} &
\begin{itemize}
    \item Tabular, 
    \item Texts, 
    \item Images and etc.
\end{itemize}
& 
\begin{itemize}
    \item Randomly shuffle labels in a small proportion of training samples;
    \item Train a model using label-modified samples and the original ones;
    \item For every training sample, collect its training loss curve per epoch and Label the curve by whether its label has been modified or not.
\end{itemize} 
& 
\begin{itemize}
    \item Train a time-series classifier of training loss curves and identify mislabeled samples through prediction.
\end{itemize} 
& 
\begin{itemize}
    \item Results of time-series classification as the likelihood of a sample being mislabeled.
\end{itemize} & 
\begin{itemize}
    \item Provides potential in enhancing performance of trained DNNs
\end{itemize} 
\\
\hline
TAPUDD~\cite{dua2023task} &
\begin{itemize}
    \item Primarily images
\end{itemize}
& 
\begin{itemize}
    \item For every sample/image from the dataset, extract the heatmaps of the sample from DNN models using local explainer tools.
\end{itemize} 
& 
\begin{itemize}
    \item Cluster the heatmap of samples in the dataset using various distance metrics.
    \item Identify outlier samples for Out-of-Distribution detection.
\end{itemize} 
& 
\begin{itemize}
    \item Outlier data points.
\end{itemize} & 
\begin{itemize}
    \item Provide task-agnostic and post-hoc OOD detection;
    \item Depend feature learning capacity of the DNN model.
\end{itemize} 
\\
\hline
PaLM~\cite{krishnan2017palm} 
&
\begin{itemize}
    \item Primarily tabular
\end{itemize}
& 
\begin{itemize}
    \item Hierarchically cluster a training dataset with a decision tree (meta-model);
    \item Partition a training dataset into subsets using intermediate nodes of the decision tree.
\end{itemize} 
& 
\begin{itemize}
    \item For every partition of training dataset, train an interpretable sub-model;
    \item Combine the decision tree (meta-model) and rules extracted from sub-models as the global explanation.
\end{itemize} 
& 
\begin{itemize}
    \item Debugging the model's response to a testing sample by iteratively tracking the logic flow.
\end{itemize} & 
\begin{itemize}
    \item Efficient in searching relevant samples;
    \item Well-balanced global and local interpretability;
    \item Difficult to scale-up on images and texts.
\end{itemize} 
\\
\hline
\end{tabular}
\label{tab:anomaly}
\end{table*}

\end{document}